\documentclass[12pt,a4paper,american]{article}

\usepackage[T1]{fontenc}
\usepackage[utf8]{inputenc}
\DeclareUnicodeCharacter{0308}{\"{} }
\usepackage{mathpazo}

\usepackage{xargs}

\usepackage{parskip}
\usepackage{babel}
\usepackage{csquotes}    
\usepackage{calc}
\usepackage[hyphens]{url}
\usepackage{makeidx}
\makeindex
\usepackage[totoc]{idxlayout}
\usepackage{booktabs} 

\usepackage{amsmath,amssymb,amsfonts}

\PassOptionsToPackage{natbib=true}{biblatex}
\usepackage[style=numeric,backend=biber,language=english,sorting=none,maxbibnames=99,doi=false]{biblatex}
\usepackage{xurl} 

\usepackage{xcolor}
\usepackage{graphicx} %

\definecolor{cvprblue}{rgb}{0.21,0.49,0.74}
\usepackage[pdfusetitle,bookmarks=true,bookmarksnumbered=true,bookmarksopen=true,bookmarksopenlevel=1,breaklinks=true,pdfborder={0 0 0},pdfborderstyle={},backref=false,colorlinks=true,allcolors=cvprblue]{hyperref}

\usepackage[capitalize]{cleveref}

\RequirePackage{xspace}
\makeatletter
\DeclareRobustCommand\onedot{\futurelet\@let@token\@onedot}
\def\@onedot{\ifx\@let@token.\else.\null\fi\xspace}

\def\eg{\emph{e.g.,}\xspace}
\def\ie{\emph{i.e}\onedot} 
 
\def\etal{et al\onedot}
\def\bda{building damage assessment\xspace}
\def\bdaAcronym{BDA\xspace}
\def\JRC{Joint Research Centre\xspace}
\def\JRCAcronym{JRC\xspace}

\usepackage{lipsum}%

\usepackage{array}
\usepackage[intoc]{nomencl}
\makenomenclature

\usepackage{a4wide}

\makeatletter

\begin{document}

\title{Revisiting the Task of Building Damage Assessment by Satellite Imaging}
\author{Anthony Cioppa, Virgile Blua, Sébastien Piérard, \\Pietro Ceccato and Marc Van Droogenbroeck}
\date{}
\maketitle
\setcounter{page}{2}

\begin{abstract}
Natural disasters such as earthquakes, tropical storms, and floods cause extensive damage to buildings, and rapid, reliable damage maps are essential to guide emergency response. Automating this task, referred to as \bda (\bdaAcronym), from satellite imagery, has become an active research topic at the intersection of remote sensing, computer vision, and machine learning, particularly since supervised deep learning methods were shown to outperform earlier handcrafted approaches. Yet, despite a decade of intense activity and the release of several large-scale datasets, the field lacks a shared foundation.

This chapter revisits the task of \bdaAcronym from first principles, organized around three pillars: data, evaluation, and methods. 
We first clarify the candidate task formulations and argue for a sensor-agnostic, instance-based definition that tolerates missing pre-event imagery. We then review existing datasets along the axes of sensing modality, temporal configuration, and geographic coverage.
On the evaluation side, we show that prevailing practices, largely based on a single aggregate score, are theoretically fragile, and we lay the foundations of a more informative, ranking-based, multi-score protocol. 
We further review methods ranging from task-specific deep networks to recent Earth observation foundation models, noticing that generalization to unseen events and regions remains an open difficulty. 
Finally, we discuss how these elements can inform the organization of a \bdaAcronym challenge, including cross-validation, challenge, and private data-sharing setups designed to build trust between developers and users.\end{abstract}

\newpage\tableofcontents{}

\global\long\def\sampleSpace{\Omega}%
\global\long\def\aSample{\omega}%
\global\long\def\eventSpace{\Sigma}%
\global\long\def\anEvent{E}%
\global\long\def\measurableSpace{(\sampleSpace,\eventSpace)}%
\global\long\def\expectedValueSymbol{\mathbf{E}}%

\global\long\def\aPerformance{P}%
\global\long\def\allPerformances{\mathbb{\aPerformance}_{\measurableSpace}}%
\global\long\def\aSetOfPerformances{\Pi}%
\global\long\def\randVarSatisfaction{S}%
\global\long\def\aScore{X}%
\global\long\def\allScores{\mathbb{\aScore}_{\measurableSpace}}%
\newcommandx\domainOfScore[1][usedefault, addprefix=\global, 1=\aScore]{\mathrm{dom}(#1)}%
\global\long\def\evaluation{\mathrm{eval}}%
\global\long\def\opFilter{\mathrm{filter}_\randVarImportance}%
\global\long\def\opNoSkill{\mathrm{no\text{\textendash{}}skill}}%
\global\long\def\opPriorShift{\mathrm{shift}_{\pi\rightarrow\pi'}}%
\global\long\def\opChangePredictedClass{\mathrm{change}_{\randVarPredictedClass}}%
\global\long\def\opChangeGroundtruthClass{\mathrm{change}_{\randVarGroundtruthClass}}%
\global\long\def\opSwapGroundtruthAndPredictedClasses{\mathrm{swap}_{\randVarGroundtruthClass\leftrightarrow\randVarPredictedClass}}%
\global\long\def\opSwapClasses{\mathrm{swap}_{\classNeg\leftrightarrow\classPos}}%
\global\long\def\allWorstPerformances{\frownie}
\global\long\def\allBestPerformances{\smiley}

\global\long\def\randVarGroundtruthClass{Y}%
\global\long\def\randVarPredictedClass{\hat{Y}}%
\global\long\def\allClasses{\mathbb{C}}%
\global\long\def\aClass{c}%
\global\long\def\classNeg{c_-}%
\global\long\def\classPos{c_+}%
\global\long\def\sampleTN{tn}%
\global\long\def\sampleFP{fp}%
\global\long\def\sampleFN{fn}%
\global\long\def\sampleTP{tp}%
\global\long\def\eventTN{\{\sampleTN\}}%
\global\long\def\eventFP{\{\sampleFP\}}%
\global\long\def\eventFN{\{\sampleFN\}}%
\global\long\def\eventTP{\{\sampleTP\}}%
\global\long\def\scorePTN{PTN}%
\global\long\def\scorePFP{PFP}%
\global\long\def\scorePFN{PFN}%
\global\long\def\scorePTP{PTP}%
\global\long\def\scoreAccuracy{A}%
\global\long\def\scoreExpectedSatisfaction{\aScore_{\randVarSatisfaction}}%
\global\long\def\scoreTNR{TNR}%
\global\long\def\scoreFPR{FPR}%
\global\long\def\scoreTPR{TPR}%
\global\long\def\scoreFNR{FNR}%
\global\long\def\scoreNPV{NPV}%
\global\long\def\scoreFOR{FOR}%
\global\long\def\scorePPV{PPV}%
\global\long\def\scorePrecision{\scorePPV}%
\global\long\def\scoreFDR{FDR}%
\global\long\def\scoreJaccardNeg{J_-}%
\global\long\def\scoreJaccardPos{J_+}%
\global\long\def\scoreCohenKappa{\kappa}%
\global\long\def\scoreScottPi{\pi}%
\global\long\def\scoreFleissKappa{\kappa}%
\global\long\def\scoreBalancedAccuracy{BA}%
\global\long\def\scoreWeightedAccuracy{WA}%
\global\long\def\scoreYoudenJ{J_Y}
\global\long\def\scorePLR{PLR}%
\global\long\def\scoreNLR{NLR}%
\global\long\def\scoreOR{OR}%
\global\long\def\scoreSNPV{SNPV}%
\global\long\def\scoreSPPV{SPPV}%
\global\long\def\scoreACP{ACP}%
\global\long\def\scoreFOne{F_{1}}%
\global\long\def\scoreFOneOverall{F_{1}^{\mathrm{overall}}}
\newcommandx\scoreFBeta[1][usedefault, addprefix=\global, 1=\beta]{F_{#1}}%
\global\long\def\scoreFOne{\scoreFBeta[1]}%
\global\long\def\priorpos{\pi_+}%
\global\long\def\priorneg{\pi_-}%
\global\long\def\scoreBiasIndex{BI}%
\global\long\def\ratepos{\tau_+}%
\global\long\def\rateneg{\tau_-}%
\global\long\def\scoreACP{ACP}%
\global\long\def\scorePFour{P_4}%
\global\long\def\normalizedConfusionMatrix{C}%
\global\long\def\scoreConfusionMatrixDeterminant{|\normalizedConfusionMatrix|}

\global\long\def\allEntities{\mathbb{E}}%
\global\long\def\entitiesToRank{\mathbb{E}}%
\global\long\def\anEntity{\epsilon}%
\global\long\def\randVarImportance{I}%
\global\long\def\randVarCanonicalImportance{\randVarImportance_{a,b}}
\global\long\def\canonicalRankingScore{\rankingScore[\randVarCanonicalImportance]}
\newcommandx\rankingScore[1][usedefault, addprefix=\global, 1=\randVarImportance]{R_{#1}}%
\global\long\def\canonicalRankingScore{\rankingScore[\randVarImportance_{a,b}]}
\global\long\def\scoreVUT{VUT}%
\global\long\def\tileCurvePriors{\gamma_\pi}
\global\long\def\tileCurveRates{\gamma_\tau}
\global\long\def\relWorseOrEquivalent{\lesssim}%
\global\long\def\relBetterOrEquivalent{\gtrsim}%
\global\long\def\relEquivalent{\sim}%
\global\long\def\relBetter{>}%
\global\long\def\relWorse{<}%
\global\long\def\relIncomparable{\not\lesseqqgtr}%
\global\long\def\rank{\mathrm{rank}_\entitiesToRank}%
\global\long\def\ordering{\relWorseOrEquivalent}%
\global\long\def\invertedOrdering{\relBetterOrEquivalent}%

\global\long\def\LScityscapes{\ding{171}}%
\global\long\def\LSade{\ding{170}}%
\global\long\def\LSvoc{\ding{169}}%
\global\long\def\LScoco{\ding{168}}%

\global\long\def\indicatorSymbol{\mathbf{1}}%
\global\long\def\realNumbers{\mathbb{R}}%
\global\long\def\aRelation{\mathcal{R}}%
\global\long\def\achievableByCombinations{\Phi}%
\global\long\def\allConvexCombinations{\mathrm{conv}}%
\newcommand{\indep}{\perp \!\!\! \perp}

\global\long\def\cityscapes{\LScityscapes{}~Cityscapes}
\global\long\def\ade{\LSade{}~ADE20K}
\global\long\def\voc{\LSvoc{}~Pascal VOC 2012}
\global\long\def\coco{\LScoco{}~COCO-Stuff 164k}

\newcommand{\MethodDesigner}{Bernadette}
\newcommand{\Benchmarker}{Leonard}
\newcommand{\AppDeveloper}{Howard}
\newcommand{\TheoreticalAnalyst}{Sheldon}

\newcommand{\tile}{Tile\xspace}
\newcommand{\tiles}{Tiles\xspace}
\newcommand{\valueTile}{Value Tile\xspace}
\newcommand{\valueTiles}{Value Tiles\xspace}
\newcommand{\baselineTile}{Baseline Value Tile\xspace}
\newcommand{\baselineTiles}{Baseline Value Tiles\xspace}
\newcommand{\SOTATile}{State-of-the-Art Value Tile\xspace}
\newcommand{\SOTATiles}{State-of-the-Art Value Tiles\xspace}
\newcommand{\noSkillTile}{No-Skill Tile\xspace}
\newcommand{\noSkillTiles}{No-Skill Tiles\xspace}
\newcommand{\skillTile}{Relative-Skill Tile\xspace}
\newcommand{\skillTiles}{Relative-Skill Tiles\xspace}
\newcommand{\correlationTile}{Correlation Tile\xspace}
\newcommand{\correlationTiles}{Correlation Tiles\xspace}
\newcommand{\rankingTile}{Ranking Tile\xspace}
\newcommand{\rankingTiles}{Ranking Tiles\xspace}
\newcommand{\entityTile}{Entity Tile\xspace}
\newcommand{\entityTiles}{Entity Tiles\xspace}

\global\long\def\aNonSkilledPerformance{\aPerformance_{\indep}}%
\global\long\def\allNonSkilledPerformances{\mathbb{\aPerformance}^{\randVarGroundtruthClass\indep\randVarPredictedClass}_{\measurableSpace}}%

\global\long\def\allPriorFixedPerformances{\mathbb{\aPerformance}^{\priorpos}_{\measurableSpace}}%

\newcommand{\comma}{\,,}
\newcommand{\point}{\,.}

\newcommandx\unconditionalProbabilisticScore[1]{\aScore_{#1}^{U}}%
\global\long\def\formulaPTN{\unconditionalProbabilisticScore{\eventTN}}%
\global\long\def\formulaPFP{\unconditionalProbabilisticScore{\eventFP}}%
\global\long\def\formulaPFN{\unconditionalProbabilisticScore{\eventFN}}%
\global\long\def\formulaPTP{\unconditionalProbabilisticScore{\eventTP}}%
\global\long\def\formulapriorneg{\unconditionalProbabilisticScore{\{\sampleTN,\sampleFP\}}}%
\global\long\def\formulapriorpos{\unconditionalProbabilisticScore{\{\sampleFN,\sampleTP\}}}%
\global\long\def\formularateneg{\unconditionalProbabilisticScore{\{\sampleTN,\sampleFN\}}}%
\global\long\def\formularatepos{\unconditionalProbabilisticScore{\{\sampleFP,\sampleTP\}}}%
\global\long\def\formulaAccuracy{\unconditionalProbabilisticScore{\{\sampleTN,\sampleTP\}}}%

\newcommandx\conditionalProbabilisticScore[2]{\aScore_{#1 \vert #2}^{C}}%
\global\long\def\formulaTNR{\conditionalProbabilisticScore{\{\sampleTN\}}{\{\sampleTN,\sampleFP\}}}%
\global\long\def\formulaTPR{\conditionalProbabilisticScore{\{\sampleTP\}}{\{\sampleFN,\sampleTP\}}}%
\global\long\def\formulaNPV{\conditionalProbabilisticScore{\{\sampleTN\}}{\{\sampleTN,\sampleFN\}}}%
\global\long\def\formulaPPV{\conditionalProbabilisticScore{\{\sampleTP\}}{\{\sampleFP,\sampleTP\}}}%
\global\long\def\formulaJaccardNeg{\conditionalProbabilisticScore{\{\sampleTN\}}{\{\sampleTN,\sampleFP,\sampleFN\}}}%
\global\long\def\formulaJaccardPos{\conditionalProbabilisticScore{\{\sampleTP\}}{\{\sampleFP,\sampleFN,\sampleTP\}}}%

\global\long\def\scoreBennettS{S_B}

\newcommand{\iquote}[1]{\textit{``#1''}}

\nomenclature{AUROC}{The \textit{Area Under the Receiver Operating Characteristic (AUROC or AUC)}\index{AUROC} is a single-number summary of a binary classifier’s ability to discriminate between the positive and negative classes. It is the probability that a randomly chosen positive instance is ranked higher (\ie, assigned a higher predicted score) than a randomly chosen negative instance by the classifier. An AUROC equal to 0.5 implies no better discrimination than random chance.}  

\nomenclature{AOI}{Acronym for \textit{area of interest}\index{AOI}. It is the geographic area where the emergency or mapping analysis is carried out, and it can be a polygon, rectangle, circle, or other defined shape.}

\nomenclature{BDA}{Acronym for \textit{building damage assessment}.}

\nomenclature{Binary}{A \textit{binary} classification task is a supervised machine learning problem where each input must be assigned to one of exactly two possible classes, such as positive vs. negative or 0 vs. 1. It is also called a crisp two-class classification task.}

\nomenclature{Classification}{A \textit{classification}\index{classification} task in machine learning is a predictive problem where the goal is to assign each input example to one label from a predefined set of categories.}

\nomenclature{Cross-validation}{\textit{Cross-validation}\index{cross-validation} is a way to evaluate how well a model will perform on new, unseen data by repeatedly training and testing it on different splits of the same dataset. It is commonly used in machine learning to get a more reliable estimate of performance and to reduce overfitting.}

\nomenclature{Data splitting}{\textit{Data splitting}\index{split} in machine learning involves dividing a dataset into distinct subsets, typically training\index{training}, validation\index{validation}, and test\index{test} sets, to train models and evaluate their performance on unseen data. This process ensures models generalize beyond the training data rather than overfitting.}

\nomenclature{Detection}{A \textit{detection}\index{detection} task in machine learning is a prediction problem where the goal is to decide whether one or more target patterns are present and, often, where they occur in the input data. It focuses on identifying the existence of specific events, objects, or anomalies rather than assigning every input to a fixed set of classes. From a more formal point of view, detection differs from binary classification in that detection does not involve true negatives.}

\nomenclature{DEM}{A \textit{Digital Elevation Model (DEM)}\index{DEM} represents the bare-earth surface, removing all natural (trees for example) and built features (building, bridges, etc.). It represents the true topographical ground surface using data, typically from LiDAR or satellite imagery, processed to remove above-ground objects.}

\nomenclature{DSM}{A \textit{A Digital Surface Model (DSM)} captures both the natural and built/artificial features of the environment.}

\nomenclature{DTM}{A \textit{Digital Terrain Model (DTM)}\index{DTM} in remote sensing is a digital representation of the terrain or ground surface of a specific area or location. It augments a a DEM in the sense that it provides information about the elevation, slope, and other topographic features of the terrain.}

\nomenclature{EMS}{EMS\index{EMS} stands for Copernicus \textit{emergency management service}. It is an EU Earth-observation service that provides geospatial information and maps to support response to natural disasters, man-made emergencies, and humanitarian crises.}

\nomenclature{FABDEM}{FABDEM\index{FABDEM} stands for ``Forest And Buildings removed Copernicus DEM''. 
It is a global bare-earth DTM derived from the Copernicus DEM. Developed at the University of Bristol and published around 2022, FABDEM addresses a well-known limitation of widely used global elevation products like SRTM and the Copernicus DEM: because those are DSMs, they systematically overestimate terrain elevation in forested and urban areas, which introduces significant errors in flood modeling, sea-level rise assessments, and hydrological routing. FABDEM uses machine learning corrections to produce a cleaner bare-earth surface, making it much closer to a true DTM at global scale.}

\nomenclature{Foundation model}{A \textit{foundation model}\index{foundation model} is a large-scale model pretrained on broad data in a task-agnostic way (usually by self-supervised learning) that can be adapted to many downstream tasks via fine-tuning or prompting; large language and multimodal models are typical examples.}

\nomenclature{Generalization}{\textit{Generalization}\index{generalization} in machine learning is the ability of a model to perform well on new, unseen data after being trained on a given dataset. It reflects how well the patterns learned from the training data apply to examples drawn from the same distribution as that data.}

\nomenclature{Imbalance}{\index{imbalance}It refers to the distribution of classes in the input data. A dataset that contains the identical amount of samples per class is said to be balanced.}

\nomenclature{MAE}{\index{MAE}Acronym for \textit{Masked Autoencoder}. It is a self-supervised approach for pretraining vision models in which a large fraction of image patches (typically ~75\%) is randomly masked and an encoder-decoder network is trained to reconstruct the missing pixels. The encoder processes only the visible patches, making pretraining efficient; the lightweight decoder is discarded afterward, leaving an encoder whose learned representations transfer well to downstream tasks.}

\nomenclature{mIoU}{The \textit{Mean Intersection over Union}\index{mIoU} calculates the Intersection over Union (IoU) for each class---defined as the overlap area between predicted and ground truth segmentation masks divided by their union area---then averages these IoU scores across all classes. }

\nomenclature{Overfitting}{\textit{Overfitting}\index{overfitting} in machine learning occurs when a model learns the training data too closely, including its noise and outliers, rather than capturing the underlying patterns. This results in excellent performance on training data but poor generalization to new, unseen data.}

\nomenclature{UAV}{\textit{UAV}\index{UAV} stands for unmanned aerial vehicle. UAVs can be terrestrial or aerial.}

\nomenclature{UAS}{An \textit{Unmanned Aerial System} (UAS)\index{UAS} is the complete operational system that includes the UAV, ground control station, communications links, payloads, data links, software, and the human operator or ground crew needed to plan, monitor, and control flights.}

\newpage
\section{Introduction}
\label{sec:introduction}

Natural disasters, such as earthquakes, tropical storms, or floods, cause extensive damage to the built environment and put populations at immediate risk. 
In the first hours following such an event, emergency responders and civil protection authorities need to know which areas are affected, which buildings are damaged or destroyed, and where relief efforts should be prioritized. 
Satellite imaging plays a central role in delivering situational awareness. 
Rapid mapping services such as the Copernicus Emergency Management Service (CEMS) On-Demand Mapping component task the acquisition of Earth observation satellite data over the affected area and produce damage maps within a few hours. 
Today, these maps are still largely produced by human experts through visual image interpretation, a process that is slow, costly, and difficult to scale when several disasters occur simultaneously. 
Automating the assessment of building damage, a task referred to as \bda (\bdaAcronym) hereafter, has therefore become an active research topic at the intersection of remote sensing, computer vision, and machine learning.

\begin{figure}[t]
    \centering
    \includegraphics[width=0.99\linewidth,alt={Qualitative result on 
    an xBD test tile of size 1024 x 1024 pixels (Hurricane Michael, 2018)}]{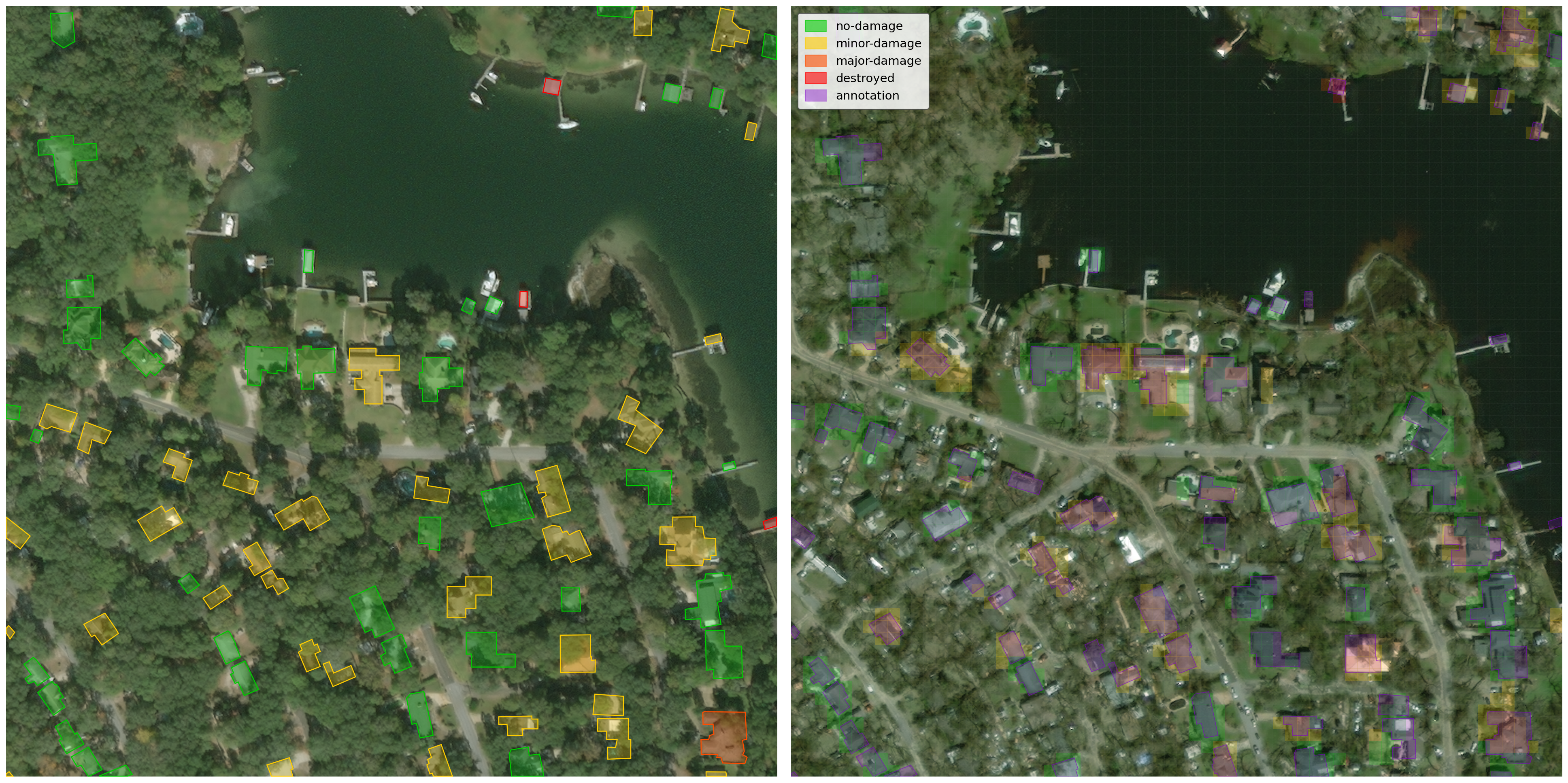}
    \caption{Qualitative result on 
    an xBD test tile of size 1024$\times$1024 pixels (Hurricane Michael, 2018). 
    Left: pre-disaster image with ground-truth footprints colored by damage class (green = no-damage, yellow = minor, orange = major, red = destroyed). Right: post-disaster image with per-patch predictions on a 64$\times$64 grid of 16$\times$16 pixel cells in the same color code;  annotations are in purple.}
    \label{fig:graphical_abstract}
\end{figure}

There are several methods for carrying out a computer vision task. 
In the past, most methods operated without learning from annotated data, \ie in an unsupervised mode. 
But over the past decade or so, thanks to its outstanding performance and exploratory research~\cite{Vetrivel2018Disaster}, supervised deep learning has become widely adopted in \bdaAcronym (\cref{fig:graphical_abstract} shows a typical result in \bdaAcronym). 
This gain in performance explains why, today, the use of data and annotations is mandatory for learning methods. 
It is all the more important to have a good understanding of the data and to ensure that the annotations are consistent and correct.

Despite a decade of intense activity and the release of several large-scale datasets, the field of \bdaAcronym suffers from a fundamental weakness: the task of \bdaAcronym itself is defined in multiple, often inconsistent, ways across the literature. 
Depending on the work, it is cast as a binary classification, a multi-class classification, semantic segmentation, or an instance-level detection problem, and these formulations imply different data requirements, different methods, and different evaluation protocols. 
This chapter therefore revisits the task of \bda from first principles, structured around three pillars: the \textit{data} (and their annotations), the \textit{evaluation}, and the \textit{methods}.

The remainder of this chapter is structured as follows.
\Cref{sec:Definition_of_tasks} examines the task of \bda: it reviews the definitions found in the literature, highlights their inconsistencies, and discusses the main candidate formulations together with their implications.
The next three sections are organized according to our three pillars. 
\Cref{sec:data} covers the data pillar: the sensing modalities, the temporal configurations, the geographic coverage, the annotation practices, and the main datasets. 
\Cref{sec:evaluation} covers the evaluation pillar: it reviews current benchmarking practices, points out their theoretical shortcomings, and lays the foundations of a standardized and informative evaluation protocol. 
\Cref{sec:methods} covers the methods pillar, from task-specific deep learning models to the recent wave of Earth observation foundation models.
Finally, \cref{sec:challenge} discusses how a challenge on \bdaAcronym could be organized, including the data splits, the evaluation strategy, and the organizational scenarios, before \cref{sec:conclusion} concludes the chapter.

\newpage
\section{Towards the definition of the task of building damage assessment}\label{sec:Definition_of_tasks}

\subsection{Definitions according to the literature}

The task of building damage assessment \index{BDA}(abbreviated as \bdaAcronym hereafter), inscribed in the earth observation literature~\cite{Janga2023AReview, Mai2025Towards, Tuia2025Artificial} is defined in multiple, and often inconsistent, ways across prior work~\cite{AlShafian2024Integrating,Gu2024Advances,Paheding2024Advancing,Raj2025AIAndGenerative}. 
The absence of a clear problem statement leading to such ambiguity in the task formulation is unfortunately common in the broader fields of image and video processing and complicates both methodological comparison and reproducibility of the results. This section reviews several definitions of the building damage assessment task proposed in the literature and highlights their limitations.

\paragraph{Lack of a problem statement or vague formulation.}
As mentioned, the lack of a precise and shared task definition is a recurring issue in the building damage assessment literature. A representative example can be found in the review by \citet{AlShafian2024Integrating}. Although the authors provide a broad overview of post-disaster building damage assessment methods, the underlying task itself is never rigorously formalized and remains conceptually vague. The authors describe it as ``\textit{This critical task of identifying damaged structures is essential for ensuring public safety, as it informs residents about the condition of their homes and supports decisions on whether they can safely reoccupy their living spaces}''. 
Yet, this formulation does not clearly state what kind of prediction is being made. It is unclear whether identifying refers to detecting buildings, segmenting them at the pixel level, performing binary or multi-class classification, estimating damage severity, or something else entirely. As will be discussed, these tasks are fundamentally different and rely on different annotations, models, and evaluation procedures. Using such a broad and informal term therefore creates ambiguity, makes comparisons between methods difficult, and blurs the practical goal of the proposed approaches.

Another example of an ill-defined task appears in \citet{Russo2026ADeep}, where the authors state: ``\textit{Following ~\citet{Sun2024QuickQuakeBuildings}, we formulate the damage detection task as an anomaly detection problem aimed at identifying buildings that exhibit structural changes in the immediate aftermath of an earthquake. While this task falls within the broader scope of \bdaAcronym, our focus is specifically on binary classification, that is, determining whether a building is damaged or intact}''. These two sentences reveal a confusion of several distinct concepts: a detection\index{detection} task, an anomaly detection task, the broader building damage assessment (\bdaAcronym) task, and a binary (two-class) classification
\index{classification} task. These formulations are not interchangeable and correspond to fundamentally different problem settings, model outputs, annotation requirements, and evaluation protocols. Ultimately, the method described by the authors performs binary classification of building damage, and the references to detection and anomaly detection serve only to introduce conceptual ambiguity rather than clarify the task.
Such inconsistencies in the task definition are problematic because they obscure the true objective of the model and prevent fair comparison across studies. A well-defined task must specify the inputs and the prediction target so that dataset construction, model design, and evaluation criteria can be aligned. 

\paragraph{\bdaAcronym as a binary anomaly detection task.}
\citet{Sun2024QuickQuakeBuildings} formulate damaged building assessment as ``\textit{a binary image classification problem, which can also be treated as an anomaly detection problem due to extreme class imbalance}''.\index{imbalance}
In a binary \index{binary} classification setting, the task consists of assigning each input sample to one of two mutually exclusive classes, here typically \textit{undamaged} (normal) versus \textit{damaged} (anomaly), based on its visual characteristics. However, within the computer vision and artificial intelligence literature, an anomaly is more precisely defined as an outlier that deviates from the underlying data distribution learned from nominal samples~\cite{Chandola2009Anomaly}. While some anomaly detection approaches rely on supervised learning, the majority are designed for semi-supervised or fully unsupervised settings, where only normal data are assumed to be available during training. In the context of building damage assessment, it is debatable whether damaged buildings truly qualify as anomalies under this definition. Although damaged samples may be statistically less frequent, their distribution is fundamentally shaped by the disaster itself and does not simply represent rare deviations from a stable, pre-existing distribution. Instead, the event induces a distribution shift, altering both the visual appearance and structural characteristics of buildings within the affected region. As a consequence, the core assumption underlying anomaly detection, namely the presence of a dominant, stationary ``normal'' distribution with sparse outliers, may not hold. This observation calls into question the suitability of framing post-disaster building damage assessment as an anomaly detection problem, even in the presence of strong class imbalance. Moreover, in cases of severe or widespread destruction, damaged buildings may become as frequent as, or even more frequent than, undamaged ones. In such scenarios, \textit{framing the task of \bdaAcronym as anomaly detection becomes particularly inappropriate}, as the damaged class can no longer be considered rare.
For instance, some works such as the one of ~\citet{Fujita2017Damage} start from 7 classes, but then binarize the problem into \textit{washed-away} versus \textit{surviving}, but still report almost $40\%$ of damaged buildings.

\paragraph{\bdaAcronym as a multi-class classification task.}
\citet{Adriano2019MultiSource} formulate building damage assessment as a multi-class classification problem using four damage levels: \textit{Destroyed}, \textit{Damaged}, \textit{Possibly Damaged}, and \textit{No Damage}, following the guidelines of the \JRC (\JRCAcronym) of the Copernicus Programme~\cite{Copernicus2018EMSR317}. In this formulation, each building is ultimately assigned to one of the four discrete damage categories. However, the proposed methodology does not operate directly at the building level. Instead, the approach first performs a per-pixel classification, effectively casting the problem as a semantic segmentation task. The resulting pixel-level predictions are then spatially aggregated over building footprints obtained from OpenStreetMap~\cite{OpenStreetMap} to derive a final label for each building. This pipeline highlights a clear mismatch between the stated task definition and the implemented methodology: although the objective is building-level damage classification, the model is trained and evaluated at the pixel level, with building-level predictions obtained only through a post-processing step. A similar four-class damage taxonomy is also adopted by the xBD dataset and benchmark~\cite{Gupta2019CreatingxBD}, one of the most widely used references in building damage assessment, with \textit{Destroyed}, \textit{Major Damage}, \textit{Minor Damage}, and \textit{No Damage}.  Hence, the changed class definition prevents direct comparison across datasets and does not resolve the underlying ambiguity between pixel-level and building-level formulations. 

\subsection{Formulation of the task of building damage assessment to automate the process}

The definition of a building damage assessment (\bdaAcronym) task must be aligned with its operational objectives. Much of the existing literature defines the task in narrow, data-specific settings, often assuming a particular sensor modality, spatial resolution, or temporal configuration. In practice, however, such assumptions are unrealistic. Disasters are unpredictable, and it cannot be guaranteed which sensing modality, for instance optical satellite imagery, Synthetic Aperture Radar (SAR), aerial data, or UAV observations, will be available immediately after an event, nor whether pre-event imagery will exist at all.

For this reason, we argue that a practical and operational task definition should be data-agnostic. The formulation should not assume access to a specific sensor nor rely systematically on pre-event imagery. Instead, it should account for heterogeneous and potentially incomplete inputs, reflecting the conditions encountered in real emergency response scenarios. In this context, the expected output of a \bdaAcronym system should be defined in a flexible yet unambiguous manner, allowing it to adapt to different downstream needs, such as rapid situational awareness, prioritization of field inspections, or integration into mapping and reporting workflows.

Importantly, the way the task is formulated directly constrains the types of methods that can be applied, the annotations that must be produced, and the evaluation protocols that can be meaningfully used. 
In the following, we examine different possible task formulations, summarized in \cref{tab:formulation_overview}, and discuss their respective strengths and limitations. 
\begin{table}[t]
    \centering
    \begin{tabular}{>{\raggedright\arraybackslash}p{0.25\linewidth}>{\centering\arraybackslash}p{0.17\linewidth}>{\raggedright\arraybackslash}p{0.48\linewidth}}
        \toprule
         \textbf{Formulation}&  \textbf{Semantic level}& \textbf{Description}\\\midrule
         Detection and binary classification&  image crop& Predict whether the crop contains a damaged building (positive class) or not.\\
         \midrule
         Multi-class classification&  image crop& Multi-class classification extends the binary setting by assigning
each image crop to one of several damage categories.\\
        \midrule
         Semantic segmentation&  pixel-level & Assigns each pixel to one of several semantic classes, such as background, undamaged building, damaged building, or destroyed building.\\ 
        \midrule
        Detection and instance segmentation& pixel-level &Localize all buildings using bounding boxes, polygons, or any shapes, and assign each instance a damage label.\\
        \bottomrule
    \end{tabular}
    \caption{Characteristics of  main possible formulations for the task of building damage assessment.}
    \label{tab:formulation_overview}
\end{table}

\paragraph{Detection and binary classification.}
A first, simple formulation consists in defining \bdaAcronym as a detection or binary classification task. In this setting, given an image crop (or a set of crops) extracted from a single (or multiple) larger remote sensing image(s), the objective is to predict whether the crop contains a damaged building or not. The subtle difference between detection and binary classification is the importance given to true negatives  (no damaged building). Binary classification is attractive due to its simplicity and relatively low annotation cost, and it is commonly used in the literature.

However, this formulation suffers from important limitations. First, it confuses multiple semantic cases into a single negative class: an image crop may contain an undamaged building, no building at all, or a partially visible structure. These situations are semantically distinct but indistinguishable under a binary label. Second, binary classification provides limited situational awareness for emergency response. 
Knowing only whether damage is present does not inform whether buildings still exist in a given area, which is critical for accessibility assessment, rescue planning, and resource allocation.

One possible extension is to decompose the problem into multiple binary tasks, for example, by first detecting the presence of buildings and then assessing their damage state. However, this naturally leads to hierarchical or multi-class formulations, which are more appropriately addressed within a unified framework.

\paragraph{Multi-class classification.}
Multi-class classification extends the binary setting by assigning each image crop to one of several damage categories, for instance: \textit{No Damage}, \textit{Possibly Damaged}, \textit{Damaged}, and \textit{Destroyed}~\cite{Copernicus2018EMSR317}. A similar formulation, with a different taxonomy, is used, for example, in the xBD dataset and benchmark~\cite{Gupta2019CreatingxBD}. A fifth class indicating the absence of buildings may also be added. 

Compared to binary classification, this formulation provides richer information and aligns more closely with operational needs. However, it introduces new challenges. The boundaries between damage classes, especially between \textit{Possibly Damaged} and other categories, are often ambiguous and difficult to annotate reliably from remote sensing data alone. In practice, the \textit{Possibly Damaged} class reflects uncertainty rather than a well-defined physical state and should not be treated on equal footing with other classes during training or evaluation. For instance, \citet{Chen2025Bright} suggest adopting a conservative approach and reclassifying them as intact.
The aspects related to annotations are further discussed in \cref{sec:data_annotations}.

Furthermore, like binary classification, this formulation relies on predefined image crops. This requires selecting a crop size and resolution a priori and raises issues when buildings are partially visible, truncated at image borders, or densely packed. As a result, localization remains implicit and imperfect.

\paragraph{Semantic segmentation.}
Semantic segmentation formulates \bdaAcronym as a pixel-level classification task. Given an image, each pixel is assigned to one of several semantic classes, such as background, undamaged building, damaged building, or destroyed building. This formulation naturally supports precise spatial localization of damage, and avoids the need for predefined image crops.

However, semantic segmentation does not distinguish between individual building instances. Adjacent buildings belonging to the same class are merged into a single connected region, making it impossible to reason at the building level without additional post-processing. Moreover, pixel-level annotations are costly and time-consuming to produce, especially at large scale and across multiple damage categories.

\paragraph{Detection and instance segmentation.}
Object detection offers a compromise between localization accuracy and annotation cost. In this setting, given a full image, the objective is to localize all buildings using bounding boxes and assign each instance a damage label. Detection naturally supports building-level reasoning, avoids cropping issues, and allows the use of existing building footprint data such as OpenStreetMap~\cite{OpenStreetMap} to pre-annotate object locations. Bounding boxes may be axis-aligned or oriented to better match building geometry, and damage labels can be assigned at the instance level. In fact, as noted by \citet{Gupta2021RescueNet}, the task of \bdaAcronym can be defined as a dual task, namely a joint building segmentation and damage assessment, illustrating the fact that there are two overlapping tasks.

If finer spatial detail is required, instance segmentation extends detection by predicting a pixel-accurate mask for each building instance. This provides more precise geometry at the cost of more expensive annotations.

Finally, panoptic segmentation combines instance segmentation for countable objects (\eg buildings, bridges) with semantic segmentation for background classes (\eg roads, vegetation). This formulation enables a unified representation of the entire scene and supports open-vocabulary extensions to additional infrastructure classes beyond buildings.

\subsection{Takeaway messages, perspectives, and open questions}

\paragraph{Implications for benchmarking.}
Each task formulation implies different data requirements, annotation strategies, and evaluation protocols. Consequently, the choice of formulation should be driven by the intended use case. A benchmark aimed at advancing scientific understanding may prioritize methodological diversity and fine-grained annotations, whereas a benchmark designed to evaluate operational readiness or select a service provider must emphasize the practical relevance of the extracted information.

In general, we argue that task definitions should remain sensor-agnostic, tolerate missing pre-event data, and support heterogeneous inputs. Semantic segmentation, detection, and instance-based formulations appear particularly well-suited to these requirements, as they balance localization, annotation feasibility, and operational interpretability.

Another important aspect relates to data availability in real-world emergency scenarios. While research datasets often assume access to high-resolution optical imagery, paired pre- and post-event observations, or carefully curated annotations, operational conditions are more variable. It is therefore important to characterize which types of data are typically available shortly after a disaster, which are only occasionally accessible, and which remain difficult to obtain. Clarifying these aspects is essential to ensuring that task definitions remain aligned with realistic operational conditions.

Finally, there is an open question regarding the nature of future benchmarks and challenges. One option is to organize scientifically oriented challenges aimed at advancing methods and driving engagement with the research community, typically over extended time frames of several months. Another option is to design more operational benchmarks focused on evaluating potential service providers. In this latter case, it remains to be determined whether the evaluation should focus solely on the performance of fully automated AI systems or instead assess the overall quality of service delivery, including the integration of limited manual corrections within constrained time budgets. Clarifying these objectives is crucial, as they directly influence task formulation, evaluation criteria, and benchmark design.

\paragraph{How can we break this deadlock?}
One strategy, adopted for example in computer vision challenges, is to define \textit{scenarios} that mimic the situations for which solutions are being sought. A scenario can specify the type of input data, the constraints on the methods to be validated, and the performance evaluation protocol. The more precise it is, the closer it will be to the real situation. On the other hand, it will be less general, and the conclusions will be applicable to a more limited number of situations. It is therefore all a matter of compromise.

\newpage
\section{Datasets for building damage assessment by satellite imaging}\label{sec:data}

Data constitute the first of our three pillars. 
Since modern methods for \bdaAcronym are data-driven, the availability, nature, and quality of the data, and above all of their annotations, condition everything that follows: the methods that can be trained and the evaluations that can be conducted. 
This section is organized in three parts. We first analyze the intrinsic characteristics along which \bdaAcronym data differ, namely the sensing modality, the temporal configuration, and the geographic scope (\cref{sec:data_axes}). We then discuss annotation practices and their practical pitfalls (\cref{sec:data_annotations}). 
Finally, we review the main datasets proposed in the literature (\cref{sec:dataset}).

\subsection{Axes of differentiation based on data}\label{sec:data_axes}

We propose to analyze the issue of data availability and type along three axes, briefly described hereafter: 

\begin{description}
    \item[Modality axis.] Datasets used for building damage assessment in the literature differ along several key axes that relate to the nature of the data itself. A first axis of differentiation concerns the data modality, that is, the type of sensor used to capture the observations. Common modalities include optical satellite imagery and Synthetic Aperture Radar (SAR), but also extend to other sources such as LiDAR, multispectral, and hyperspectral imagery. In addition to traditional remote sensing data, some works also leverage alternative modalities, including social media content and external building maps.

    \item[Temporal axis.] A second important axis relates to temporality, namely whether datasets rely on single-temporal data (typically post-event imagery only) or multi-temporal data that include both pre-event and post-event observations. This distinction has significant implications for the types of methods that can be applied, particularly regarding change detection versus post-event-only analysis.
    
    \item[Contextual axis.] A third axis concerns the spatial and contextual scope of the data. This includes the geographic regions covered by the datasets, as well as the types of disasters represented within those regions. Differences in building styles, urban density, environmental conditions, and disaster characteristics can strongly influence dataset composition and model generalization.
\end{description}

These axes of differentiation are intrinsic to the data and are independent of the specific task formulations discussed in the previous section. In principle, the various task formulations, \ie, classification, detection, or segmentation, can all be instantiated using different combinations of these data characteristics. In the following, we describe each data type axis in more detail and discuss their respective implications for building damage assessment. 

Note that questions about annotations relating to the state of buildings for a \bdaAcronym task play a special role. These annotations are useful either for developing methods or models, or for evaluating the performance of a method. As we will see later, some datasets provide annotations of this type (see \cref{sec:data_annotations}).

\subsubsection{Modalities}
\label{sec:data_modalities}

Researchers and practitioners rely on a wide range of sensing modalities to assess building damage following disasters. These include traditional remote sensing technologies such as optical satellite imagery, Synthetic Aperture Radar\index{SAR} (SAR), and LiDAR,\index{LiDAR} as well as non-traditional data sources such as social media content and building maps. Each modality provides different types of information and comes with distinct advantages and limitations in terms of spatial resolution, temporal availability, robustness to environmental conditions, and ease of integration into operational workflows. This section reviews these modalities and discusses their respective roles in building damage assessment.

\paragraph{Optical imagery satellites.} 
Optical satellite imagery is one of the most widely used data sources for building damage assessment, offering broad geographic coverage and high spatial resolution. Public missions such as Copernicus Sentinel-2\index{Sentinel-2} provide multispectral imagery at a spatial resolution of up to 10~m with a frequent revisit time of approximately five days, enabling large-scale monitoring. Commercial satellites, such as WorldView-4, offer very high-resolution (VHR)\index{VHR} imagery, with spatial resolutions reaching up to 0.3~m, allowing for more detailed analysis of individual structures.
The advantage of having such high resolution is that we can use terrain models that account for building heights. However, this advantage can become a problem when the viewing angle between pre- and post-images differs in the case of analyzing damaged buildings.

Despite showing some advantages, optical imagery exhibits important limitations for damage assessment. As noted by \citet{Adriano2019MultiSource}, optical data can reliably capture severely affected structures, such as collapsed buildings following earthquakes or washed-away buildings after tsunami events. However, distinguishing intermediate levels of structural damage remains challenging. This limitation stems from the predominantly nadir, overhead viewpoint of satellite optical imagery, which restricts visibility of façades and other critical structural elements. Consequently, indicators of partial damage, such as façade cracking, internal collapse, or localized deformation, are often not observable or may be obscured by rooftops, debris, or shadows. As a result, the level of detail achievable for damage characterization using optical satellite imagery alone remains limited, particularly for subtle or moderate damage states.

\paragraph{Synthetic aperture radar satellites.}
Synthetic Aperture Radar (SAR)\index{SAR} systems emit microwave signals that interact with the Earth's surface and record the backscattered echoes to produce high-resolution images. These measurements provide valuable information about surface roughness, structural changes, and ground deformation, all of which are relevant indicators of potential building damage in post-disaster scenarios.
A wide range of SAR satellite missions have been used for damage assessment, including TerraSAR-X, TanDEM-X, Sentinel-1\index{Sentinel-1}, RADARSAT-2, ALOS-2, RISAT, COSMO-SkyMed, and Gaofen-3. In addition to spaceborne sensors, some studies have explored airborne SAR acquisitions to obtain very high-resolution imagery for detailed damage mapping~\cite{Shi2015Building}.
One of the main advantages of SAR imagery in disaster response contexts is its independence from illumination and weather conditions. SAR sensors can operate through cloud cover, fog, rain, and snow, and do not rely on daylight, enabling continuous 24/7 ground imaging. This capability is particularly valuable for rapid post-event assessment, when optical imagery may be unavailable for extended periods.
From a physical perspective, layover scattering patterns observed in SAR intensity images can reveal information about site-wall interactions, providing cues about the structural condition of buildings after a disaster. Furthermore, different backscattering mechanisms captured through multi-polarization SAR data offer complementary information that can help characterize various damage states~\cite{Adriano2019MultiSource}.
Despite these advantages, SAR imagery also presents inherent challenges. SAR data are affected by radiometric and geometric distortions, Doppler-related artifacts, and speckle noise~\cite{AlShafian2024Integrating}, which can complicate both visual interpretation and automated analysis. These artifacts may obscure fine structural details, reduce image contrast, and introduce ambiguities in object boundaries, thereby limiting the accuracy of building delineation and damage severity estimation when SAR data are used in isolation.

\paragraph{UAV.}\index{UAV} Airborne remote sensing using unmanned aerial vehicles (UAVs) enables the acquisition of very high-resolution imagery, often reaching spatial resolutions on the order of a few centimeters. Compared to satellite-based observations, UAV platforms can support multi-view imaging, with sensors capturing the scene from different viewing angles. This capability provides valuable information for assessing building façades and other structural elements that are largely invisible from nadir-only satellite perspectives. 

Beyond still imagery, UAVs can also acquire video data, potentially from multiple viewpoints. Such multi-view video streams enable temporal analysis of damage evolution and can support the monitoring of progressive structural degradation. These observations may be further enriched by integrating complementary video sources, such as security cameras or other available live streams, thereby contributing to a more continuous and comprehensive situational overview in post-disaster contexts. In addition to visual sensing, UAVs can be equipped with microphones, enabling the capture of acoustic signals. In specific scenarios, such audio data may convey valuable information, for instance, through the detection of human voices, collapse sounds, secondary structural failures, or other acoustic signatures indicative of ongoing hazards or survivability conditions.

UAV\index{UAV} platforms can also carry LiDAR\index{LiDAR} (Light Detection and Ranging) sensors, enabling the generation of dense three-dimensional point clouds that provide an additional geometric modality for post-disaster analysis. LiDAR systems operate reliably in low-light and nighttime conditions and are less affected by certain atmospheric factors. However, deploying LiDAR-equipped UAVs remains costly, and point clouds present inherent limitations: they cannot penetrate buildings, and their spatial sampling density often prevents the detection of fine-grained surface cracks.
Despite these advantages, UAV-based data acquisition is subject to important operational constraints. Flight duration is limited by battery capacity, which can restrict coverage in remote or large-scale disaster areas. UAVs are also sensitive to adverse weather conditions. Furthermore, aviation regulations and airspace restrictions often require that UAV operations be conducted by authorized local entities, limiting the flexibility and immediacy of deployment in emergency response scenarios. Finally, the use of UAVs can lead to serious translational alignment errors, as studied by \citet{Manzini2025NonUniform}.

\paragraph{Social media data.}
More recently, researchers have begun to explore unconventional data modalities derived from social media\index{social media} as indirect sources for estimating physical phenomena related to the Earth~\cite{Raj2025AIAndGenerative}. In particular, \citet{Mousavi2024Gemini} demonstrated that multimodal, unstructured social media posts can be leveraged to estimate earthquake ground-shaking intensity, highlighting the potential of crowd-sourced information for rapid disaster assessment.
These findings open promising avenues for building damage assessment using social media content. Such approaches may exploit geolocated textual information as well as image geolocalization techniques enabled by recent advances in computer vision~\cite{Berton2023EigenPlaces, Berton2022Rethinking,Berton2022Deep, Mereu2022Learning}. For instance, recent methods such as MegaLoc~\cite{Berton2025MegaLoc} demonstrate that place recognition remains feasible even under severe domain shifts, including changes in illumination (day versus night) and weather conditions (rain versus sunny scenes). Extending this line of research to train models capable of localizing images under substantially more challenging conditions, such as scenes containing partially or fully destroyed buildings, represents a largely unexplored direction. Such capabilities would be particularly valuable in post-disaster contexts, where reference imagery may no longer reflect the current state of the environment. Robust geolocalization of citizen-generated content could therefore play a role in rapidly integrating social media data into building damage assessment pipelines, complementing traditional remote sensing observations.

\paragraph{Other types of modalities.}
In addition to imagery and video data, auxiliary information sources are often leveraged to support building damage assessment. A prominent example is OpenStreetMap\index{OpenStreetMap} (OSM)~\cite{OpenStreetMap}, which provides worldwide geographic information, including building footprints. These footprints are frequently used in the construction of building-damage datasets to indicate the approximate location and spatial extent of structures~\cite{Adriano2019MultiSource}. In practice, such annotations allow researchers to convert pixel-level segmentation problems into building-level classification tasks by extracting image crops around buildings known to exist before the disaster.
However, OSM data are generated and maintained mainly by volunteers, leading to variability in completeness, accuracy, and temporal consistency across regions. As a result, building footprints may contain errors, omissions, or outdated information, particularly in areas experiencing rapid urban development or limited mapping activity. These limitations introduce uncertainty into both dataset construction and model evaluation. In some works, OSM building footprints are not only used for annotation purposes but are also directly incorporated as inputs to the proposed methods~\cite{Russo2026ADeep}, further amplifying the impact of potential inaccuracies on downstream performance.

\subsubsection{Single vs. multi-temporality}

Datasets used for building damage assessment are commonly categorized according to their temporal structure. They are either (1) single-temporal, relying solely on post-disaster imagery, or (2) multi-temporal, providing paired pre- and post-event data to enable change-detection-based analysis. This distinction is important, as the temporal configuration of a dataset directly influences the design, assumptions, and operational scope of damage assessment models.

As highlighted in the survey by \citet{Gu2024Advances}, methods based on change detection face significant practical limitations. In many developing regions, reliable and consistent pre-disaster remote sensing data are unavailable or incomplete, which restricts the applicability of approaches that rely on temporal image pairs. Even when pre-event imagery exists, it may be outdated. Buildings constructed between the time of image acquisition and the disaster are absent from the baseline data, leading to missed detections or false change signals caused by urban development rather than disaster-induced damage.

Because disasters are inherently unpredictable, ensuring access to recent and high-quality pre-event imagery for any given region is not feasible. For these reasons, operational building damage assessment systems should primarily rely on post-event observations, treating pre-event data as optional supplementary information when it is available, rather than as a strict requirement.

\subsubsection{Studied world regions}

The geographic scope of datasets plays an important role in the development and evaluation of building damage assessment methods. Differences in architectural styles, urban layouts, construction materials, and environmental conditions across regions strongly influence both visual appearance and damage patterns, making geographic diversity a key factor for model robustness. In particular, geographic generalization\index{generalization} remains a major challenge in building damage assessment. In their survey, \citet{Gu2024Advances} note that none of the reviewed approaches, whether based on classical algorithmic techniques or modern deep learning methods, demonstrate strong robustness across different regions. This highlights a persistent gap between real-world performance and the ones reported in benchmarks.

Some studies attempt to address this issue by analyzing the impact of a single disaster event across multiple locations. For instance, \citet{Russo2026ADeep} investigate the effects of the 2023 T\"urkiye earthquake on several cities located around the epicenter. To assess generalization, the authors adopt a leave-one-out strategy, training on a subset of cities and testing on a held-out one. While this approach is an improvement over training and testing on the same city, it remains confined to a geographically and structurally homogeneous context. As a result, it provides limited insight into how the proposed methods would perform in entirely different regions characterized by distinct architectural styles, urban layouts, building materials, or environmental conditions. Such constraints limit the conclusions that can be drawn regarding deployment in diverse, real-world scenarios.

At a larger scale, the xBD\index{xBD} dataset~\cite{Gupta2019CreatingxBD} includes imagery from 11 major disaster events, along with 8 additional events, spanning the period from 2011 to 2019. While this geographic and temporal diversity represents a significant step forward, the official documentation does not clearly specify whether the same regions may appear across multiple dataset splits (training, validation, and test). This ambiguity raises concerns about potential spatial leakage, which could artificially inflate reported performance and hinder a fair assessment of a model’s ability to generalize to unseen regions or future disaster events.

While the xBD dataset~\cite{Gupta2019CreatingxBD} represents an important step in the right direction by incorporating multiple disasters and geographic regions, a large portion of the literature still focuses on a single disaster event occurring within a single region. Such experimental settings substantially limit the ability to assess model robustness and often lead to overly optimistic performance estimates. Addressing this limitation requires datasets that deliberately span diverse regions and disaster types, coupled with evaluation protocols designed to test cross-region generalization. One effective strategy consists in splitting\index{split} datasets into training, validation, test, and potentially challenge subsets, such that each split contains distinct geographic regions, with any given region-disaster pair appearing in only one split. This type of geographically disjoint partitioning is a well-established approach to reduce overfitting\index{overfitting} and to more accurately evaluate a model’s ability to generalize to unseen regions and future disaster scenarios.

\subsection{Annotations}\label{sec:data_annotations}
\index{annotation}

\subsubsection{The annotation practices}

A pragmatic evaluation approach consists in defining a ground truth and comparing oneself to it. With that respect, a reliable proposal for annotations was provided by the \citet{IWG-SEM2018Emergency}\index{IWG-SEM}. They proposed a four-level granularity scale for building damages: (1) \textit{destroyed}, (2) \textit{damaged}, (3) \textit{possibly damaged}, and (4) \textit{no visible damage}, with precise definitions and visual examples to help annotators. 

Likewise, in operational CEMS On-Demand Mapping (ODM) practice, the annotations of building damages typically correspond to point locations with a legend restricted to (i)~\textit{Destroyed}, (ii)~\textit{Damaged}, and (iii)~\textit{Possibly Damaged}, produced manually using visual image interpretation.

If we model the task of \bdaAcronym as a multi-class classification task, we can consider that there is a fourth category called \textit{other}. However, this can pose problems because it is not possible to know whether the pixels or objects in this class are intact, whether they are background, or whether the corresponding area has not been annotated.

In reality, the annotation practice is useful as it provides a broad outline of the objectives to be achieved by giving the output to be predicted. Yet, this approach does not specify either the application scenario or the scores to be used for the evaluation. For example, in most benchmarks, methods are evaluated by collapsing certain classes to reduce the problem to a two-class classification problem. 
Furthermore, since annotations are done by hand, they are not entirely reliable, both because there is overlap between class definitions---the ``possibly damaged'' class is subject to interpretation---and for other reasons, such as the presence of debris near buildings, misalignments, or geolocalization errors. 

There are other sets of classes for annotation that provide different nuances, as shown in \cref{tab:bright-annotations,tab:xBD-annotations}; these tables present the categories and definitions of annotated classes for the BRIGHT~\cite{Chen2025Bright} and xBD~\cite{Gupta2019CreatingxBD} datasets, respectively.

\begin{table}[thbp]
    \centering
    \caption{Definition of different categories in the BRIGHT\index{BRIGHT} dataset~\cite{Chen2025Bright}.}
    \begin{tabular}{>{\raggedright\arraybackslash}p{0.185\linewidth}>{\raggedright\arraybackslash}p{0.75\linewidth}}
    \toprule
         \textbf{Category (label)}& \textbf{Definition}\\
    \midrule
         Background (0) & All non-building pixels\\
    \midrule
         Intact (1) & No visible signs of structural damage, water intrusion, shingle displacement, or burn marks\\
    \midrule
         Damaged (2) & Partial structural damage to the building, such as missing roof members, visible cracks, or partial wall/roof collapse. Buildings may be partially burned, surrounded by water or mud, or affected by nearby volcanic flows.\\
    \midrule
         Destroyed (3) & Completely collapsed, burned, partially/completely covered by water/mud, or no longer present\\
     \bottomrule

    \end{tabular}
    \label{tab:bright-annotations}
\end{table}

\begin{table}[thbp]
    \centering
    \caption{The classes in the xBD\index{xBD} dataset follow a four-level granularity scheme~\cite{Gupta2019CreatingxBD}.}
    \begin{tabular}{>{\raggedright\arraybackslash}p{0.20\linewidth}>{\raggedright\arraybackslash}p{0.725\linewidth}}
    \toprule
         \textbf{Category (label)}& \textbf{Definition}\\
    \midrule
         No Damage (0)& Undisturbed. No sign of water, structural or shingle damage, or burn marks.\\
    \midrule
         Minor Damage (1)& Building partially burnt, water surrounding structure, volcanic flow nearby, roof elements missing, or visible cracks.\\
    \midrule
         Major Damage (2)& Partial wall or roof collapse, encroaching volcanic flow, or surrounded by water/mud.\\
    \midrule
         Destroyed (3)& Scorched, completely collapsed, partially/completely covered with water/mud, or otherwise no longer present.\\
     \bottomrule

    \end{tabular}
    \label{tab:xBD-annotations}
\end{table}

In conclusion, there are practical differences concerning data annotation, summarized in \cref{tab:annotations-summary}. 
\begin{table}[ht]
    \centering
    \caption{Typical annotation categories.}
    \begin{tabular}{c>{\centering\arraybackslash}p{0.22\linewidth}>{\centering\arraybackslash}p{0.22\linewidth}}
    \toprule
            \textbf{CEMS ODM/IWG-SEM}~\cite{IWG-SEM2018Emergency}\index{JRC}\index{IWG-SEM}&\textbf{BRIGHT}~\cite{Chen2025Bright}\index{BRIGHT}&\textbf{xBD}~\cite{Gupta2019CreatingxBD}\index{xBD} \\
    \midrule
            Destroyed&Destroyed&Destroyed  \\
    \midrule
            Damaged&Damaged&Major Damage \\
    \midrule
            Possibly damaged&Intact&Minor Damage \\
    \midrule
            (No visible damage~\cite{IWG-SEM2018Emergency})&Background&No Damage \\
     \bottomrule
    \end{tabular}
    \label{tab:annotations-summary}
\end{table} 

Regardless of the choice made, it is recommended to provide as precise a definition as possible and to give examples for the different modalities, as was done in \citet{IWG-SEM2018Emergency}. Ideally, one should also consider the evaluation strategy to align the annotations with the application objectives.

\subsubsection{Practical issues}

In practice, the quality of annotations is a major issue for training the methods, which goes hand in hand with performance evaluation. For example, errors in the selected class can occur. Indeed, the ``possibly damaged'' class is assigned when a building is suspected of being damaged, either due to the presence of debris nearby or a damaged building in the neighborhood. There are also difficulties in geolocating damaged buildings, partly due to the angles at which high-resolution images are captured. Another challenge is the type of annotation, which varies from segmentation masks to predefined shapes such as segments, bounding boxes, convex sets, or even a simple point at the center of a building. Finally, it sometimes happens that certain parts are overlooked due to a lack of time.

\subsection{Building damage assessment datasets}\label{sec:dataset}

\subsubsection{Identification of articles and descriptions of current or former datasets}

This section reviews the datasets\index{dataset} proposed in the literature for building damage assessment and summarizes their main characteristics. A comparative overview of these datasets across key dimensions, such as modality, disaster type, and annotation type, is provided in \cref{tab:dataset_overview}. As can be seen, the diversity of datasets highlights substantial variability in task definitions (binary vs. multi-class classification, segmentation, or mixed formulations), input modalities (optical, SAR, UAV, street-view, or social media imagery), and annotation strategies (image-level, pixel-level, or building-instance-level), which complicates direct comparison across methods and benchmarks. Hereafter, we provide a description of a few prominent datasets.

\begin{table}[p]
\centering
\caption{Overview of representative BDA datasets, summarizing task formulation, sensing modality, temporal setting (post-event only vs. paired pre/post), geographic coverage, disaster types, number of annotated building instances, and annotation granularity.}
\label{tab:dataset_overview}
\medskip
\noindent\resizebox{\linewidth}{!}{
\begin{tabular}{>{\raggedright\arraybackslash}p{0.15\linewidth}>{\centering\arraybackslash}p{0.15\linewidth}>{\centering\arraybackslash}p{0.15\linewidth}>{\centering\arraybackslash}p{0.14\linewidth}>{\centering\arraybackslash}p{0.10\linewidth}>{\centering\arraybackslash}p{0.16\linewidth}>{\centering\arraybackslash}p{0.1\linewidth}>{\centering\arraybackslash}p{0.21\linewidth}}
\toprule
\textbf{Dataset (year)} &
\textbf{Task} &
\textbf{Modality} &
\textbf{Temporality} &
\textbf{Region} &
\textbf{Disaster type} &
\textbf{No. of building} &
\textbf{Annotation type} \\
\midrule

ABCD~\cite{Fujita2017Damage} (\citeyear{Fujita2017Damage}) & Binary classification & Aerial image (PASCO) & Pre- and post- event & Japan & Tsunami & 22,171 & Image crop level \\
\hline
\citet{Nguyen2017Damage} (\citeyear{Nguyen2017Damage})     & Multi-class classification & Social media images & Post-event & 4 regions & Hurricane, typhoon, earthquake & 25,758 & Image-level (3 labels) \\
\hline
Cheng et al.~\cite{Cheng2021Deep} (\citeyear{Cheng2021Deep})& Multi-class classification & UAV aerial imagery & Post-event & Caribbean & Hurricane & 6,440 & Instance masks (2 labels) \\
\hline
\citet{Xue2024Posthurricane-arxiv} (\citeyear{Xue2024Posthurricane-arxiv}) &
Multi-class classification & Street-view images + struct. data & Post-event & Florida, USA & Hurricane & 2,472 & Building-level (3-5 labels) \\
\hline
FloodNet~\cite{Rahnemoonfar2021FloodNet} (\citeyear{Rahnemoonfar2021FloodNet}) &
Classification, segmentation, VQA\index{VQA} & UAV aerial imagery & Post-event & Texas, USA & Flood & 8,946 instances& Pixel-level semantic (9 labels) + VQA \\
\hline
RescueNet~\cite{Rahnemoonfar2023RescueNet} (\citeyear{Rahnemoonfar2023RescueNet}) & Classification, semantic segmentation & UAV aerial imagery & Post-event & Florida, USA &
Hurricane & 10,903 instances& Pixel-level segmentation masks + image-level (10 labels) \\
\hline
Ida-BD~\cite{Kaur2023Large-scale} (\citeyear{Kaur2023Large-scale}) & Multi-class classification, segmentation & Optical (VHR satellite) & Pre- and post-event & Louisiana, USA & Hurricane & N/A & Building-instance polygons (4 labels) \\
\hline
CRASAR-U-DROIDS~\cite{Manzini2024CRASARUDROIDs-arxiv} (\citeyear{Manzini2024CRASARUDROIDs-arxiv})& Building damage assessment & sUAS\index{UAS} orthomosaic imagery & Post-event & USA  & Multiple disaster types & 21,716 & Building-instance polygons (4-5 labels)\\
\hline
Noto-BDA-MV~\cite{Vescovo2025The2024} (\citeyear{Vescovo2025The2024}) & Binary classification & Optical + auxiliary imagery &
Pre- and post-event & Japan & Earthquake & 140,208 & Building-instance polygons \\
\hline
xBD~\cite{Gupta2019CreatingxBD} (\citeyear{Gupta2019CreatingxBD}) &
Multi-class classification & Optical satellite imagery & Pre- and post- event & Worldwide  & Multiple natural disasters & 850,736 & Building-instance polygons (4 labels) \\
\hline
xFBD~\cite{Melamed2022xFBD-arxiv} (\citeyear{Melamed2022xFBD-arxiv}) & Multi-class classification, instance segmentation & Optical satellite imagery & Pre- and post- event (synthethic composites, subset of xBD~\cite{Gupta2019CreatingxBD}) & Palu, Indonesia  & Tsunami & 8,027 & Building-instance polygons (4 labels)\\
\hline
QuickQuake\-Buildings~\cite{Sun2024QuickQuakeBuildings} (\citeyear{Sun2024QuickQuakeBuildings}) & Binary classification & Optical + SAR satellite imagery & Post-event & Islahiye, Turkey &
Earthquake & 4,029 & Building-level binary labels \\
\hline
BDD~\cite{Adriano2021Learning} (\citeyear{Adriano2021Learning}) &
Semantic segmentation  & Optical + SAR remote sensing & Pre- and post-event & Multiple regions & Earthquake, tsunami, typhoon & N/A & Pixel-level semantic masks (3 labels) \\
\hline
BRIGHT~\cite{Chen2025Bright} (\citeyear{Chen2025Bright})  & Classification, segmentation & Optical + SAR satellite imagery & Pre- and post-event & Worldwide & 7 types of disasters & 384,596 & Building-instance polygons (4 labels) \\

\bottomrule
\end{tabular}
}
\end{table}

\paragraph{xBD.} 
A first prominent example is the xBD\index{xBD} dataset~\cite{Gupta2019CreatingxBD}, whose scale and diversity have had a substantial impact on the field of building damage assessment. Distributed through the Maxar DigitalGlobe Open Data Program~\cite{Vantor2026Maxar}, xBD provides extensive pre- and post-event multiband satellite imagery covering 11 major disasters along with 8 additional events, spanning the period from 2011 to 2019. These events include earthquakes, wildfires, tornadoes, floods, and volcanic eruptions across multiple geographic regions. Building footprints and damage labels are manually annotated using polygonal outlines, followed by two rounds of quality control involving both annotators and domain experts, resulting in comparatively high-quality annotations at scale.

\paragraph{xFBD.} 
\citet{Melamed2022xFBD-arxiv} proposed the xFBD\index{xFDB} (Focused Building Damage) dataset as a second, smaller, version of xBD specifically designed to evaluate whether methods can assess damage at the individual building level without relying on surrounding context. The dataset is constructed by isolating single damaged buildings from the xBD Palu Tsunami subset and blending them into undamaged pre-event imagery using Poisson blending, producing realistic composite post-event images in which only one building exhibits damage. The xFBD dataset comprises $8{,}027$ synthetic image pairs ($5{,}287$ training, $1{,}466$ test, and $1{,}274$ holdout samples) drawn from 131 source images, with damage annotations inherited from the original xBD labels. Notably, the dataset contains almost exclusively destroyed and major damage samples, with only one minor damage sample. While xFBD is derived from a single disaster and region and is therefore limited in geographic diversity, it addresses a complementary question to standard benchmarks: whether a method's predictions are driven by building-level features or by the appearance of the surrounding area.

\paragraph{Sun \etal.} 
\citet{Sun2024QuickQuakeBuildings} introduce the first dataset designed for detecting earthquake-damaged buildings using post-event very high-resolution (VHR)\index{VHR} Synthetic Aperture Radar (SAR)\index{SAR} and optical imagery, based on data from the 2023 Turkey-Syria earthquake. The dataset comprises co-registered post-disaster SAR and optical observations covering more than $4{,}000$ building footprints. However, the authors acknowledge that aligning SAR imagery with building footprints is a non-trivial task. To address this challenge, they rely on an automatic co-registration algorithm~\cite{Sun2020Automatic} that exploits building features corresponding to the bases of sensor-visible walls present in both SAR and footprint modalities. This approach assumes that buildings remain largely upright, such that these structural features are still observable and geometrically consistent across sensors. While this assumption may be valid for lightly damaged structures, it is likely to break down in cases of severe damage or complete collapse, where the relevant geometric cues are absent or heavily distorted. The potential impact of this limitation on the quality of the co-registration and consequently on the reliability of the dataset is not discussed by the authors. They only mention manually checking the alignment.

\paragraph{Adriano \etal.} 
\citet{Adriano2019MultiSource} also leverage both optical and SAR data, combining observations from four sensors: PlanetScope, Sentinel-2, Sentinel-1, and ALOS-2 PALSAR-2. In contrast to purely post-event approaches, their method adopts a multi-temporal formulation using both pre- and post-disaster imagery from the 2018 Sulawesi earthquake and tsunami in Palu, Indonesia. However, their evaluation protocol draws both training and test samples from the same geographic area. Such a setting is poorly aligned with real-world disaster response scenarios, where models must generalize to new regions affected by future events. This experimental design risks significant overfitting and is therefore likely to overestimate performance when applied to unseen environments.

\paragraph{BRIGHT.} 
More recently, \citet{Chen2025Bright} proposed the BRIGHT dataset, a large-scale multimodal benchmark combining very high-resolution optical and SAR imagery to enable all-weather building damage assessment. %
BRIGHT covers 14 disaster events across 23 regions worldwide, spanning five types of natural disasters and two types of human-made disasters
The dataset includes paired pre- and post-event imagery and building-level damage annotations, making it one of the most geographically diverse resources currently available. By explicitly targeting multimodal and multiregional generalization, BRIGHT represents a significant step toward more operationally relevant benchmarking.

\subsection{Takeaway messages, perspectives, and open questions}

The review of existing datasets and task formulations for building damage assessment reveals both significant progress and persistent limitations in the current scientific literature. While substantial efforts have been made to collect large-scale datasets and develop increasingly sophisticated models, the lack of consensus on task definitions, data assumptions, and evaluation protocols remains a major obstacle to meaningful comparison.

A first key takeaway is the \textit{strong heterogeneity in how building damage assessment is formulated}. Across datasets and benchmarks, the task is alternately cast as binary classification, multi-class classification, semantic segmentation, or instance-level segmentation. These formulations correspond to fundamentally different objectives, input assumptions, and outputs, yet they are often discussed under a single umbrella term: building damage assessment.

Second, most \textit{existing datasets are tightly coupled to specific sensing modalities or temporal configurations}. Many approaches assume the availability of pre-event imagery, even though disasters are unpredictable and up-to-date pre-disaster data cannot be guaranteed in practice. Conversely, datasets relying exclusively on post-event data often embed strong assumptions about sensor type or spatial resolution. From an operational standpoint, and particularly from the perspective of organizations such as the Copernicus Emergency Management Service On-Demand Mapping, task definitions and benchmarks should be largely data-agnostic and robust to heterogeneous, partially missing, or asynchronous inputs.

A third important observation concerns generalization and geographic bias. Although recent datasets such as xBD and BRIGHT make clear progress toward multi-region coverage, \textit{a large fraction of studies still evaluate models on a single event or a single region}, frequently using training and test samples drawn from the same geographic area. Such protocols tend to overestimate performance and provide limited insight into how methods will behave when deployed in new environments with different building typologies, urban layouts, or imaging conditions. Geographically disjoint evaluation splits should therefore be considered a minimal requirement for future benchmarks.

Finally, these observations naturally raise several questions. From a data perspective, \textit{which modalities are commonly accessible (\eg optical satellites, SAR), which are sporadically available (\eg UAV imagery, street-level views), and which remain largely opportunistic (\eg social media content)?}
It seems that optical imagery is the most commonly used modality, alongside SAR sensors, which prove useful in the case of cloud cover or nighttime conditions, particularly for tropical storms.
Street-level and social media content seem to be underexplored currently, which could be an interesting research field to explore in the future.
Beyond availability, \emph{what levels of spatial resolution or geolocalization accuracy are required to support current mapping products and decision-making workflows?} 
Very high-resolution imagery is the most commonly available type of resolution in operational rapid mapping, but the geometric alignment issues discussed in this section suggest that higher resolution does not automatically translate into more reliable annotations.

\newpage
\section{Towards building an evaluation protocol of algorithms for building damage assessment}\label{sec:evaluation}

The previous sections have shown that both the definition of the \bda (\bdaAcronym) task and the organization of data are complex. This is also the case when it comes to evaluating methods designed to solve a \bdaAcronym task.
 
In practice, evaluation can have multiple objectives: developing a particular method to maximize an objective function relative to a known ground truth, comparing methods to select the best one in a specific situation, selecting the method that seems most appropriate for predicting damages in a new situation, or even understanding the strengths and weaknesses of each method, which may involve a degree of subjectivity.

To start the discussion on the evaluation, we will first describe current practices (see \cref{sec:evaluation_practices}). We will see that current practices are imperfect and raise a whole series of questions. Next, to lay the foundations for an appropriate strategy, we will break down the theoretical elements to be considered in an evaluation process in \cref{sec:evaluation-theory}. Finally, \cref{sec:evaluation-new} is more prospective and presents some lines of thought for defining evaluation scenarios for \bdaAcronym.

\subsection{Evaluation practices}\label{sec:evaluation_practices}

\subsubsection{Benchmarking practices}

\paragraph{The xBD's benchmark.} A prominent example is the xBD\index{xBD} dataset~\cite{Gupta2019CreatingxBD}, whose large scale and diversity have had a major impact on the field. By offering extensive pre- and post-event multi-band satellite imagery across a wide range of disasters, xBD catalyzed progress in building damage assessment. The dataset was further popularized through the xView2 Challenge, which provided a structured benchmark and motivated the development of numerous influential architectures~\cite{Weber2020Building, Zheng2021Building}.
Consequently, xBD has established itself as a reference benchmark against which new \bdaAcronym models are commonly evaluated.

\paragraph{Evaluation on the Turkey earthquake and the xView2 datasets.} In their survey, \citet{Gu2024Advances} provide a comparison of seven methods evaluated on two datasets: the 2023 Turkey earthquake and the xView2 datasets. This work is notable as one of the few attempts to benchmark both classical algorithm-based techniques as well as more modern deep learning methods within a common framework. Nevertheless, the analysis relies exclusively on qualitative visualizations and does not present any quantitative results. Based on the provided examples, it appears that all models are judged using segmentation outputs, even when some of them were originally designed for different prediction paradigms, such as object detection models like \index{YOLO} YOLOv8~\cite{Varghese2024YOLOv8}. This inconsistency highlights the absence of a clear task definition and problem statement, ultimately resulting in a weak and non-comparable benchmark. 

\paragraph{Evaluating the \bda task (\bdaAcronym) as a detection or binary classification task.} 
As noted previously, most \bdaAcronym benchmarks formulate the task as a detection or a binary classification task (see \cref{tab:benchmarking-scores}). To proceed to an objective evaluation, a large body of work on binary image classification relies on a limited set of performance scores, often referred to as metrics, although these quantities do not define metrics in the strict mathematical sense, which is a common abuse of terminology in the literature. These scores are typically derived from the confusion matrix, which comprises four components\footnote{Note that to obtain these four events, it is needed to collapse the annotations into the damaged vs. undamaged classes.}: 
\begin{itemize}
    \item True Positives (TP), corresponding to damaged buildings correctly classified as damaged; 
    \item False Positives (FP), undamaged buildings incorrectly classified as damaged; 
    \item False Negatives (FN), damaged buildings incorrectly classified as undamaged; and 
    \item True Negatives (TN), undamaged buildings correctly classified as undamaged. 
\end{itemize}
The confusion matrix is the base of many scores used in benchmarks, as can be seen in~\cref{tab:benchmarking-scores}, which also mentions the evaluation strategy and defines the scores used to evaluate methods. 

\begin{table}[t]
    \centering
    \caption{Overview of the most representative benchmarks in building damage assessment with their respective tasks, evaluation strategy, and evaluation scores. Note that the AI Model Challenge is not public.}
    \begin{tabular}{>{\raggedright\arraybackslash}p{0.17\linewidth}>{\centering\arraybackslash}p{0.155\linewidth}
    >{\centering\arraybackslash}p{0.265\linewidth}
    >{\centering\arraybackslash}p{0.29\linewidth}}
    \toprule
         \textbf{Benchmark} & \textbf{Task} & \textbf{Evaluation strategy} & \textbf{Score definition}\\
    \midrule
         xBD~\cite{Gupta2019CreatingxBD} & Multi-class classification &  Per-class $\scoreFOne$\index{$\scoreFOne$} and Weighted $\scoreFOne$ & $\scoreFOne = \frac{2TP}{2TP + FP + FN}$\\
         \midrule
         QuickQuake\-Building \cite{Sun2024QuickQuakeBuildings} & Binary classification & Precision\index{precision}, recall\index{recall}, $\scoreFOne$, and AUROC\index{AUROC} & $P=\frac{TP}{TP+FP}$, $R=\frac{TP}{TP+FN}$, $\scoreFOne=2\frac{PR}{P+R}$ (alternative equivalent definition)\\
         \midrule
         BRIGHT~\cite{Chen2025Bright} & Classification, segmentation & Overall accuracy\index{accuracy} (OA), $\scoreFOne-$score, split into $\scoreFOne^{loc}$ and $\scoreFOne^{cls}$, and mean intersection over union (mIoU)\index{IoU} & $OA = \frac{\sum TP_i}{\#elements}$ $IoU = \frac{TP}{TP + FP + FN}$ \\
         \midrule
         {AI} Model Challenge~\cite{InternationalCharter2025AIModel} & Detection (binary classification) & $\scoreFOne$ at the building level & $\scoreFOne=2\frac{PR}{P+R}$ \\
    \bottomrule
    \end{tabular}
    \label{tab:benchmarking-scores}
\end{table}

Because of a class imbalance\index{imbalance} between the positives and negatives, it is common to ignore true negatives (TNs) by considering the task as a detection task. For instance, \citet{Sun2024QuickQuakeBuildings} report only precision $P=\frac{TP}{TP+FP}$, recall $R=\frac{TP}{TP+FN}$, and the $\scoreFOne$ score. However, the $\scoreFOne$ score is fully determined by precision and recall, as it is their harmonic mean, and therefore does not provide additional information beyond those two quantities. Moreover, this limited set of scores offers only a partial view of classifier behavior and obscures the role of other error types, more precisely the role of true negatives, which can be important in damage assessment scenarios.

\paragraph{Evaluation on the QuickQuakeBuilding dataset.} \citet{Sun2024QuickQuakeBuildings} further argue that their problem is closer to an anomaly detection setting due to the strong class imbalance toward undamaged samples. On this basis, they adopt the Area Under the Receiver Operating Characteristic (AUROC or AUC)\index{AUROC} as an evaluation criterion. However, computing the ROC curve requires considering a continuum of classifiers obtained by varying the decision threshold on a confidence score (such a classifier is referred to as a \textit{soft classifier}). As a result, the AUROC does not evaluate a single operational classifier but rather an entire family of hypothetical classifiers. This disconnect limits its relevance for practical deployment, where a fixed decision rule is ultimately required.

\paragraph{Further discussion.} 
As mentioned earlier, an ambiguity arises at the level of the confusion matrix itself. In many formulations, the absence of a building is implicitly merged with the ``undamaged building'' class. This ambiguity fundamentally alters the definition of positive and negative samples and directly affects how the confusion matrix is constructed. Consequently, evaluation results are highly sensitive to the underlying task definition—specifically, whether the problem is framed as building presence detection, damage classification conditional on building existence, or a joint formulation of both. This lack of clarity in problem specification undermines the interpretability and comparability of reported performance.

Furthermore, some works such as \citet{Russo2026ADeep} refer to detection accuracy as an evaluation score%
, although this term is ill-defined in the context of building damage detection. Accuracy is a score derived from the binary confusion matrix as $A = \frac{TP + TN}{TP + FP + FN + TN}$, which includes true negative (TN) values. However, detection tasks disregard TNs, as the space of ``on-objects'' is not well defined, making accuracy an unsuitable score in such settings.

In the case of \citet{Russo2026ADeep}, the problem is ultimately framed as a binary classification task distinguishing between intact and damaged buildings, and the authors compute standard scores including precision,\index{precision} recall,\index{recall} $\scoreFOne$ score,\index{$\scoreFOne$} AUROC,\index{AUROC} and Cohen's Kappa.\index{Cohen's Kappa} 
Yet, these scores are sensitive to the arbitrary choice of which class is considered ``positive''. Since the positive/negative assignment is a priori invertible (either damaged or intact buildings could be labeled as positive), different choices lead to different score values and rankings. In this case, the intact class is treated as the negative class, seemingly because it is the majority in the dataset. However, in severely affected cities---where most buildings are damaged---, this choice could arguably be inverted, raising important questions about consistency in the evaluation process.

This highlights a broader issue: the lack of a clearly defined task and evaluation protocol can introduce ambiguity not only in model training but also in score and ranking interpretation. 
Interestingly, the qualitative results in \citet{Russo2026ADeep} are presented as map overlays where buildings are colored red or blue to indicate damage or no damage, respectively. This visualization style is more typical of segmentation tasks than of binary classification or detection, further illustrating the conceptual mismatch between the stated task and the actual methodology.

In the multi-class classification setting, studies such as the xBD\index{xBD} dataset paper~\cite{Gupta2019CreatingxBD} report evaluation scores including per-class $\scoreFOne$-score, precision, and recall. These scores are computed separately for each damage class, without any form of aggregation across classes (neither \textit{micro}- nor \textit{macro}-averaging), even though the authors mention a ``weighted $\scoreFOne$ score'', without providing further details. The authors argue that accuracy is an inadequate score due to the strong class imbalance, with approximately 75\% of the samples belonging to a single category. The authors mention that the accuracy is a flawed score due to the high imbalance (with around 75\% for one class), but it is arguable that this limitation is also valid for any defined score and that $\scoreFOne$, precision, and recall do not solve for these imbalance issues. It seems that this choice is more based on a, rather cautious, per-class evaluation strategy rather than on known practices in multi-class classification.  

Finally, benchmarks that formulate building damage assessment as a semantic segmentation task, such as BRIGHT~\cite{Chen2025Bright}, primarily rely on segmentation-specific evaluation scores, most notably the mean Intersection over Union\index{mIoU} (mIoU). However, these works also report classical classification scores, such as overall accuracy, which corresponds to the accuracy computed from the induced multi-class confusion matrix. Moreover, BRIGHT introduces a decomposition of the $\scoreFOne$ score into $\scoreFOne^{\text{loc}}$ and $\scoreFOne^{\text{cls}}$, intended to separately evaluate localization and classification performance. However, the precise definitions of these scores, including how localization is quantified, and how these scores are aggregated across classes and instances, are not explicitly specified. This lack of formal definition makes the reported results difficult to interpret.

\subsection{Theoretical elements of an evaluation procedure}\label{sec:evaluation-theory}

In this section, we discuss some theoretical aspects related to evaluation. These elements will be considered later on to define an evaluation protocol or guide certain decisions with a view to setting up a challenge.

\paragraph{Underlying statistical assumption.}
\citet{Japkowicz2011Evaluating} presents several ways of approaching the issue of evaluation. On the one hand, there is the \textit{statistical approach}, which consists in determining the statistical significance of the results. This approach is rarely used in computer vision because we usually assume that we have enough data available—we could legitimately ask whether this statement holds true for the \bda task—but also because there are other sources of error during the annotation process. On the other hand, there is the \textit{probabilistic approach} based on the frequentist hypothesis~\citet{Neyman1977Frequentist} of probabilities. This is the approach commonly used in the field of \bdaAcronym. Added to this are underlying assumptions of spatial stationarity of observations and ergodicity. 

\paragraph{On the use of scores.} 
The use of scores can have three different purposes: (1) optimizing the parameters of a method during its development, (2) comparing methods or variants of a method, and (3) comparing a set of methods to select the best one. 

For the development of a method (point (1)), particularization to a score leads to an evaluation bias. In our practice, we also noted that if a single score is widely adopted in a community, such as the $\scoreFOne$ score in \bdaAcronym, methods tend to resemble each other and have very similar performances. This effect, called the \textit{attractor effect} by \citet{Zakharova2026Event-arxiv}, was observed in the AI Model Challenge for Earthquake Response \cite{InternationalCharter2025AIModel} in these terms:  \textit{because of the closeness of results and the quality of the submissions, the challenge organizers and Charter partners decided to recognize the top four for this challenge}. The solution is to use a multi-criteria approach to avoid any evaluation bias. 

From a practical perspective, comparing methods (point (2)) or selecting a method (point (3)) can be done using the same ranking process. The theoretical question then is whether a score is adequate for ranking, as indicated by \citet{Pierard2025Foundations}. This is the case for the $\scoreFOne$ score or the IoU\footnote{Most developers are unaware that the $\scoreFOne$ score and the IoU provide the same ranking and present perfectly correlated lists without noticing it, as in \cite{Bou2025ChangeDetection}.}, but this is not valid anymore if the scores are averaged, as in the case of mIoU, unless the contextual average is used \cite{Pierard2020Summarizing}. 

\paragraph{Crisp vs. soft classifiers.} 
The disadvantage of modeling the task of \bdaAcronym as a binary classification is that a model developed for this purpose is designed to determine a positive or negative with certainty; it is then not possible to qualify the confidence of a prediction. A soft classifier associates a confidence score with a prediction, allowing a user to adjust the expected result. The disadvantage is then determining which threshold to choose to provide the optimal result. Another problem is the fact that it is difficult to compare a crisp classifier and a soft classifier.

\paragraph{Generalization and transferability.}
Several benchmarks consider distinct geographic areas to enable multi-domain evaluation, adopting a leave-one-out strategy. The idea is to evaluate the generalization capabilities of a method, which is a motivation for the development of so-called \textit{foundation models}. However, as noted for multi-domain adaptability~\cite{Pierard2025Multidomain-arxiv} or model transferability, the underlying theoretical issues are largely ignored. 

The use of the $\scoreFOne$ score can lead to unexpected difficulties. For example, in the International Contest on Illegal Waste Dumping Detection (IWDD) challenge, which adopted the $\scoreFOne$ score for evaluation (see \url{https://mivia.unisa.it/iwddcontest2026/}), it was found that among all the methods submitted, only one method outperformed a no-skill method (\ie, a method producing a completely random output) in terms of accuracy on the challenge set, while all methods performed acceptably on the test set. In fact, no method obtained a $\scoreFOne$ score higher than a no-skill classifier on the challenge set.

When considering ranking, another difficulty is that the ranking obtained on an evaluation set is not necessarily indicative of a method's rank for other data, as noted by \citet{Pierard2025AMethodology}. Overall, predicting the rank of a method for a new dataset is a problem that, to our knowledge, has not been studied in the scientific literature.
\subsection{Towards a standardized and informative evaluation protocol: takeaway messages and perspectives}\label{sec:evaluation-new}

Performance evaluation in the field of building damage assessment is often based on a single score, whether it is the $\scoreFOne$ score, mIoU, accuracy, or any other criterion. This form of evaluation has several shortcomings: 

\begin{enumerate}
    
    \item Using a score to compare and rank methods requires that the score used satisfy a very specific axiom, as established by \citet{Pierard2025Foundations}. Thus, not all scores are ranking scores, and not under all conditions (some require working with constant priors, which is difficult to guarantee for new images).
 
    \item  Interpreting the absolute value of a score only makes sense when approaching the upper limit, but remains insignificant for values in the middle of the scale. Furthermore, some arguments that are valid for the value of a score are not transferable when it comes to ranking. Thus, \citet{Flach2015Precision} have shown that while the $\scoreFOne$ score is the harmonic mean between precision and recall, it is not necessarily the best compromise between these two scores when evaluating a soft classifier by thresholding. Worse still, \citet{Pierard2026What} have established that, in many scenarios, the ranking induced by the $\scoreFOne$ score is more strongly correlated with the rankings induced by precision or recall, and therefore does not induce the best intermediate ranking.
 
    \item A single criterion does not express the full potential of a method. As shown in \cite{Pierard2025AMethodology}, there is currently no reliable method for predicting the rank of a method on a new dataset based on ranks obtained under different conditions.   
   
    \item Adopting a single score impoverishes the development of new methods. Indeed, it encourages the development of similar methods with incremental rather than disruptive improvements.
    
    \item Adopting a single score carries the risk of overfitting on the score rather than on the task.
 \end{enumerate}

The approach we propose is based on important principles:

\begin{enumerate}
    \item Evaluation must emphasize the comparison of methods through the concept of \textit{ranking}; score values are one factor among others, but the stability of the ranking must take precedence to arrive at lasting conclusions.

    \item A \textit{multiple-score approach} must be adopted. This is a way to better understand the potential of a method. A tool that may be suitable when the task is similar to that of a crisp classifier is the concept of \textbf{Tile}\index{Tile} presented in~\cite{Pierard2024TheTile-arxiv}.

    \item The evaluation must consider \textit{different distinct application cases}. For example, sometimes the objective is to have the best ratio of correct detections, while other times the goal is to minimize false positives, etc. In all cases, the definition of application constraints must remain the prerogative of the end user of a method.

    \item Ideally, a theoretically sound strategy is needed for \textit{multi-domain evaluation}.
    
    \item The evaluation must potentially allow for an estimation of performance in a \textit{transferability} mode to anticipate performance in a new situation.

\end{enumerate}

Finally, a major concern in choosing a method is the level of \textit{trust} a \textit{user} can place in the solution proposed by a \textit{developer} (vendor).
The trust between a developer and a user is partly based on the performance evaluation process. However, as explained in this section, using a single score to evaluate performance and compare methods is not a best practice. For instance, the $\scoreFOne$ score is widely favored for evaluating methods in \bda, just as it is by the scientific community in computer vision ($10\%$ of the papers at the main CVPR conference mention or use the $\scoreFOne$ score!), which is far from ideal. The reasons cited for choosing the $\scoreFOne$ score are that it is a good tradeoff between precision and recall, since it is defined as the harmonic mean of these two values. This idea is contested by \citet{Flach2003TheGeometry}, who proposes a skew-insensitive variant of the $\scoreFOne$ score. Very recently, \citet{Pierard2026What} showed that neither $\scoreFOne$ nor its skew-insensitive version are acceptable tradeoffs for \emph{ranking}, in general. Indeed, typically, there is a confusion between the fact that a value is a tradeoff and the fact that the derived ranking is the optimal tradeoff among the rankings of the induced values. The reality is that the $\scoreFOne$ score is suboptimal for ranking and that it is better to choose the $F_\beta$ score, with a value of $\beta$ that depends on the distribution of the analyzed performances.

Beyond the choice of scores, trust also depends on how data and annotations are shared between the user and the developer. This question, namely the comparison of the cross-validation, challenge, and private data-sharing setups, and how trust evolves over time in each of them, is analyzed in detail in \cref{sec:challenge}.

\newpage
\section{Methods}
\label{sec:methods}

Recently, neural networks have naturally emerged as the primary approach used to assess damage to buildings. They allow for the processing of large volumes of data, as well as the classification and localization of damage directly on satellite images. The techniques and architectures used vary considerably and depend on the task to be accomplished and the quality of the data.

\subsection{The open-source literature}
\label{sec:method-Literature}

The literature presents a highly heterogeneous landscape~\cite{Ahn2025Generalizable, Chachra2022Detecting, Chen2024CDasXORNet, Chen2025Bright, Dietrich2025AnOpensource, MorenoGonzalez2025Toward, Hafner2025DisasterAdaptiveNet, Li2025BuildingGuided, Lv2025Earthquake}. Articles combine proprietary and open-source data, mixing multimodal optical imagery, drone imagery, ground-level photographs, or subsets of existing datasets, rather than relying on standardized benchmarks. 
This fragmentation actually complicates the establishment of reliable performance benchmarks for competing algorithms. As pointed in \cref{sec:dataset}, among the open-source datasets, two stand out: xBD~\cite{Gupta2019CreatingxBD} and BRIGHT~\cite{Chen2025Bright}.

In the case of the xBD, the data was officially split
into the ``train'' (``tier1'' + ``tier3'') set, the ``test'' set, and the ``holdout'' set.
The ``holdout'' set was kept confidential throughout the challenge and used
exclusively by the organizers to evaluate the generalization performance of
submitted models, while ``tier3'' was released midway through the competition
as supplemental training data covering additional hazard types and geographic
areas. The baseline model was trained on a ResNet-50~\cite{He2016DeepResidual} architecture with dense output layers for classification. The official ranking score was
$\scoreFOneOverall = 0.3 \, \scoreFOne^{\mathrm{loc}} + 0.7 \, \scoreFOne^{\mathrm{dmg}}$,
where $\scoreFOne^{\mathrm{loc}}$ scores building localization and $\scoreFOne^{\mathrm{dmg}}$ is the harmonic mean of the four per-class damage $\scoreFOne$ scores (no damage,
minor, major and destroyed).
As the authors point out, for highly imbalanced datasets, accuracy alone is an imperfect score: a classifier that systematically predicts ``no damage'' on all images would maintain 75\% accuracy, masking its total inability to detect actual damage. The baseline model achieved only an overall weighted $\scoreFOne$ score of 0.2654, revealing the difficulty of the problem of damage classification in real-world conditions. This poor performance can be explained by several factors: the massive class imbalance (with ``no damage'' representing more than 75\% of the samples), the subtle visual differences between ``minor damage'' and ``major damage'' that cause confusion between these classes, and the intentional labeling noise introduced to manage the ambiguity inherent in damage assessment. These results demonstrate that despite a rigorous annotation process and expert quality control, the task of generalization from limited and imbalanced training data remains a major challenge.

In BRIGHT, the standard training configuration combines pre-event optical imagery and post-event SAR imagery, enabling damage assessment across all conditions. Additional tests show that performance varies depending on the type of post-disaster imagery used. With SAR alone, the average mIoU reaches 65.6\%, while it rises to 69.8\% with optical images (pre-event optical + post-event optical). BRIGHT's paper~\cite{Chen2025Bright} shows that optical+SAR fusion (integrating pre-event and post-event data from both modalities) yields the best results with a mIoU score of 70.8\%, confirming their complementarity. Optical imaging provides fine visual details, while SAR provides structural information that is robust to weather and smoke. However, the inter-event generalization still remains challenging. Indeed, in ``zero-shot'' transfer, where a new disaster is analyzed without any labeled data from that event, the average performance drops to less than 40\% mIoU, compared to around 50\% for fully supervised training\footnote{The concept of supervised learning represents the use of labeled data where each input has a corresponding ground-truth label, and usually requires extensive manual annotation. In contrast, the self-supervised training approach learns from unlabeled data by creating pseudo-tasks (like predicting masked parts of images), eliminating the need for human labels. Semi-supervised training combines both approaches, using a small amount of labeled data alongside a larger pool of unlabeled data to improve performance while reducing annotation costs.}.

The ChangeMamba model retains the best robustness with 39.4\% in zero-shot. To reduce this gap, the authors evaluate semi-supervised learning (SSL) approaches that exploit a few annotated examples of a new disaster, a practice known as ``few-shot''. These techniques yield gains of up to 4 to 6 mIoU points over zero-shot (43.2\% in one-shot for ChangeMamba), illustrating their potential to improve transferability in real-world conditions where labels are scarce.

Other datasets merging optical and SAR imagery exist, such as QQB~\cite{Sun2024QuickQuakeBuildings} and BDD~\cite{Adriano2021Learning}. However, QQB covers only a single event with 4,029 annotations and BRIGHT improves on BDD on three fronts: finer spatial resolution for both modalities (0.3–1 m vs. 1.2 m), broader and more diverse disaster coverage, and open-source availability.

As data sprawls, so do the tasks. Among benchmarked tasks, localization and classification remain the most prevalent. This disparity in datasets naturally extends to the modeling approaches used. Historically, researchers developed task-specific models trained from scratch or adapted general computer vision architectures, resulting in limited generalizability and high computational costs for each new application. To meet these challenges, so-called ``foundation'' models have emerged, featuring architectures designed to meet various evaluation requirements.

\subsection{Foundation models}

This paradigm shift has been driven by the massive amounts of data provided by satellite constellations, which have fundamentally transformed the field. Over the past two years, dozens of foundation models have emerged to leverage this wealth of data, for example: SpectralGPT~\cite{Hong2024SpectralGPT}, DOFA~\cite{Xiong2024Neural-arxiv}/Clay~\cite{Clay2024TheFoundationModel}, Copernicus-FM~\cite{Wang2025TowardsAUnified}, Galileo~\cite{Tseng2025Galileo}, Prithvi/EO-2.0~\cite{Jakubik2023Foundation-arxiv, Szwarcman2026PrithviEO2.0}, TerraMind~\cite{Jakubik2025TerraMind}, and more recently AlphaEarth~\cite{Brown2025AlphaEarth-arxiv}. 
Frameworks like TerraTorch~\cite{Gomes2025TerraTorch} and TorchGeo~\cite{Stewart2025TorchGeo} to use and build around these models also start to appear. 
A ``foundation'' model is a deep learning model trained on massive data volumes, designed for versatility and adaptability across multiple tasks through fine-tuning or prompting. Typical examples include GPT~\cite{Radford2018GPT}, BERT~\cite{Devlin2019BERT}, CLIP~\cite{Radford2021Learning}, and SAM~\cite{Kirillov2023Segment} in computer vision. 

Based on this, \citet{Rotich2025Evaluating} assessed the approaches of eight foundation models. All models rely on Vision Transformer (ViT~\cite{Dosovitskiy2021AnImage}) architectures except GFM~\cite{Mendieta2023Towards}, which uses Swin-B~\cite{Liu2021Swin}. The objective is to understand how these models, pre-trained on large geospatial corpora, perform across diverse downstream tasks such as classification, oriented object detection, building segmentation, and change detection. 
The authors chose 4 self-supervised methods, pure MAE~\cite{He2022Masked} and three MAE-based approaches: SatMAE~\cite{Cong2022SatMAE}, Scale-MAE~\cite{Reed2023ScaleMAE}, and Cross-Scale MAE~\cite{Tang2023CrossScale}, another SSL variant called the GFM~\cite{Mendieta2023Towards}, and finally a supervised approach called Multi-Task Pretraining~\cite{Wang2024MTP} (MTP). Each model was pre-trained on one of these four databases: fMoW~\cite{Christie2018Functional}, SAMRS~\cite{Wang2023SAMRS}, Million-AID~\cite{Long2021OnCreating}, and GeoPile~\cite{Mendieta2023Towards}. The models were then fine-tuned on three target datasets: DOTA-v2.0~\cite{Xia2018DOTA} for oriented object detection, BigEarthNet~\cite{Sumbul2019BigEarthNet} for multi-label classification, and xBD for segmentation and change detection.

The predominance of MAE-based approaches reflects a broader strategy in remote sensing to address the scarcity and cost of labeled data. The MAE method consists of masking random areas of input images and reconstructing the missing pixels, enabling models to learn rich visual representations. This self-supervised approach is particularly valuable in remote sensing where data labeling is costly, time-consuming, and requires domain expertise, yet massive volumes of raw satellite data are readily available. A classic two-phase training paradigm was used:  (1) during the initial unsupervised pre-training phase on massive data volumes, models assimilate fundamental visual structures and semantic relationships without explicit labels; (2) the subsequent fine-tuning phase then specializes these models for specific applications using limited labeled examples. This approach allows models to capture the essence of satellite imagery characteristics before any formal supervision, making the final learning more efficient and generalizable across diverse remote sensing tasks.

The performance of the models built is then compared with the state-of-the-art model results on the three different datasets selected to evaluate the tasks. It is also important to note that beyond the four pre-training datasets used in this study (fMoW, SAMRS, Million-AID and GeoPile), numerous other foundation datasets exist in the remote sensing domain, including SSL4EO-S12~\cite{Wang2023SSL4EOS12}, Copernicus-Pretrain~\cite{Wang2025TowardsAUnified}, RS5M/GeoRSCLIP~\cite{Zhang2024RS5M}, MMEarth~\cite{Nedungadi2024MMEarth}, SeeFar~\cite{Lowman2024SeeFar-arxiv}, M3LEO~\cite{Allen2024M3LEO} and TerraMesh~\cite{Blumenstiel2025TerraMesh}. Each of them offers different characteristics in terms of scale, diversity and coverage. The choice of a pre-training dataset is a critical factor that shapes the capabilities of the resulting model and the performance of the downstream task~\cite{Rotich2025Evaluating}.

For the xBD dataset in the change detection category, the best performance (an $\scoreFOneOverall$ 
score of 77.62) is achieved with the MTP method using a ViTL+RVSA~\cite{Wang2024MTP} backbone pre-trained on the SAMRS dataset. In the building segmentation category, the ScaleMAE ViTL architecture on the fMoW dataset achieves an $\scoreFOneOverall$ score of 77.83.

Performance on xBD is evaluated against other models from the literature trained on ImageNet~\cite{Deng2009ImageNet}, including Siamese-UNet~\cite{CayeDaudt2018Fully} (ResNet34), ChangeOS~\cite{Zheng2021Building} (ResNet50), DamFormer~\cite{Chen2022DualTasks} (ViT-B), MambaBDA~\cite{Chen2024ChangeMamba} variants (ViM-T, ViM-S, ViM-B) based on ViM~\cite{Zhu2024Vision}, and BDANet~\cite{Shen2022BDANet} (ViT-B). The best overall model is MambaBDA (ViM-B) with an $\scoreFOneOverall$ score of 81.84. Across most models, localization scores are notably higher than damage level classification scores, with BDANet being the exception.
These results reveal a fundamental pattern of task dependency in foundation model performance. While foundation models match or exceed supervised baselines on image classification and segmentation tasks, they struggle significantly with change detection and oriented object detection. This pattern is consistent across all evaluated datasets: models fail to reach 80\% on change detection (except MambaBDA+ViT-B) and 60\% on oriented object detection (except RHINO~\cite{Lee2025Hausdorff} at 60.62 mAP).

Several key factors emerge from this analysis. Dataset spectral and spatial diversity proves more critical than sheer data volume for effective pre-training. Naively increasing model size does not yield proportional performance improvements, highlighting the need for optimized task-agnostic architectures. Vision Transformers slightly outperform CNNs when pre-trained on relevant geospatial data. Additionally, semi-supervised learning techniques demonstrate that algorithms using unlabeled training data can achieve performance comparable to fully supervised algorithms trained on labeled datasets, offering a practical path forward given the scarcity of labeled remote sensing data.

Five years ago, the SKAI~\cite{Lee2020Assessing-arxiv} model release by Google Research already anticipated these findings and introduced one of the first applications of semi-supervised learning to building damage assessment using minimal labeled data alongside large amounts of unlabeled data. The approach applies two semi-supervised learning techniques: MixMatch~\cite{Berthelot2019MixMatch} and FixMatch~\cite{Sohn2020FixMatch}. They both create augmented images and calculate loss based on prediction quality. For ground-truth labels of undamaged buildings, the authors used a pretrained building detection model to identify structures and filtered out those marked as damaged in UNOSAT assessments, resulting in a two-class classification scheme: ``damaged'' and ``undamaged''.

Results demonstrated well that these semi-supervised models achieve performance close to fully supervised learning using only a fraction of annotations, validated across three disasters: the Santa Rosa wildfire, Haiti earthquakes, and Aleppo snapshot. SKAI compares these approaches against the Twin Tower~\cite{Xu2019Building-arxiv} model and a fully supervised baseline trained on all labeled data, confirming again that SSL techniques can match fully supervised performance with significantly less labeled data. However, the study also already revealed a well-known limitation, quote: ``\textit{[...] models trained only on past disasters will likely underperform in new disasters}'', highlighting the fundamental generalization challenges also noted years ago by  \citet{Valentijn2020MultiHazard}.

SKAI's approach reveals also persistent biases that semi-supervised learning alone cannot eliminate. The systematic centering of crops on detected buildings actually can introduce spatial bias, where models may learn positional regularities rather than actual damage characteristics. Without validation on truly novel disasters, generalization remains unproven, echoing the broader challenge identified by \citet{Torralba2011Unbiased} over a decade ago: even massive, diverse datasets contain inherent biases that compromise model generalization.

Three years later, a work by \citet{Yeh2023Debiased-arxiv} on the xView~\cite{Lam2018xView-arxiv} dataset corroborated these findings while revealing additional bias sources inherent to semi-supervised methods. Their adapted FixMatch approach achieves 82.4\% accuracy using only 30\% labeled data. However, they identify pseudo-label imbalance as a critical issue: SSL frameworks generate biased predictions that reinforce majority classes, creating false majorities during training. Their debiasing approach addresses both human-labeled class imbalance and model-generated pseudo-label bias, demonstrating that effective SSL for remote sensing requires careful attention to multiple bias sources beyond spatial positioning.

A META research team~\cite{Liu2025ADecade} demonstrated recently this phenomenon by showing that modern neural networks excel at capturing dataset-specific patterns, achieving 84.7\% accuracy in classifying image origins between YFCC~\cite{Thomee2016YFCC100M}, CC~\cite{Changpinyo2021Conceptual}, and DataComp~\cite{Gadre2023DataComp} datasets. These results suggest models capture dataset-specific signatures that can mask their true ability to generalize. For satellite imagery, such biases are inherent due to heterogeneous acquisition parameters across different sensors and satellites, as noted in the AlphaEarth paper~\cite{Brown2025AlphaEarth-arxiv}. 

Additionally, there are surveys covering more broadly foundation models as well as their limitations and benchmarks in remote sensing that have been published recently~\cite{Huang2025ASurvey-arxiv, Huo2025When, Xiao2025Foundation, Zhu2026OnThe}.

\subsection{The MAE revolution}

The fact is that Masked AutoEncoder~\cite{He2022Masked} has become the dominant paradigm to build earth observation foundation models. To understand how this revolution enabled a leap in performances, it seems important to grasp how it actually works and what its creation brought about. The method works by randomly masking a percentage (usually 75\%, but possibly up to 98.5\%~\cite{Eymael2024Efficient}) of image patches, encoding the remaining patches with a Vision Transformer (ViT), then reconstructing the masked patches through a lightweight decoder while minimizing mean squared error (MSE) between reconstructed and original pixels.

MAE's success stems from three fundamental advantages over previous approaches. First, it achieves extreme simplicity: a single augmentation technique (random masking) suffices, unlike earlier contrastive methods such as SimCLR~\cite{Chen2020ASimple} and MoCo~\cite{He2020Momentum} for example, that required sophisticated augmentations and very large batch sizes (4096+). Second, it delivers exceptional computational efficiency by encoding only 25\% of patches, reducing computational cost by a factor of three. Third, it demonstrates scalability with predictable scaling laws where performance improves continuously with more data, larger models, and longer training, following principles similar to GPT language models.

These advantages translate into impressive empirical results. On ImageNet, this purely self-supervised approach achieves 87.8\% top-1 accuracy after fine-tuning, rivaling traditional supervised methods while being considerably simpler to implement and train than previous self-supervised techniques. Implicitly, the adoption of MAE in earth observation is directly linked to the specific challenges of this field. The fundamental problem in earth observation stays the massive imbalance between data and labels: satellite missions generate tens of petabytes of imagery (Sentinel-2: 5+ PB since 2015, Landsat: 50+ years of archives, Sentinel-1: 3+ PB of radar data) while ground-truth labels remain generally scarce (LCMAP CONUS: ~27k points, GLaNCE global: ~35k points), creating a imbalanced data/label ratio. MAE solves this problem by enabling exploitation of all unlabeled data through self-supervision, where each image serves as its supervision.

It then naturally exploits the massive spatial redundancy present in satellite imagery. A typical Sentinel-2 image of a homogeneous forest can contain 70\% nearly identical pixels, meaning that masking 75\% of patches does not eliminate 75\% of information thanks to this redundancy, unlike ImageNet natural images where each patch may represent a different object. The masking forces the model to learn multi-scale patterns essential in earth observation: at local scale (10-100m) it learns texture and spatial continuity, at regional scale (1-10km) it captures spatial context (agricultural vs. urban zones), and at landscape scale (10-100km) it extracts global ecological patterns (biomes, climatic signatures).

The MAE approach also adapts naturally to the multi-spectral character of satellite images without architectural modifications. While Sentinel-2 has 13 spectral bands (visible, near-infrared, and shortwave infrared) and Landsat has 11 bands (including thermal), masking works identically for all bands, avoiding the thorny problem of defining relevant data augmentations for non-visual channels. 

More remarkably, MAE offers natural robustness to clouds, a common issue in earth observation. If we say clouds naturally mask 20-40\% of a typical satellite image, a model pre-trained with 75\% random masking has actually already learned to ``hallucinate'' missing patch content, conferring operational robustness to cloud contamination. Finally, MAE outputs (spatially dense feature maps typically 128$\times$128$\times$768) align perfectly with dominant downstream tasks such as: semantic segmentation for land cover mapping, object detection for building identification, change detection, and biophysical variable regression (biomass, evapotranspiration), all requiring representations preserving spatial structure rather than global embeddings.

This alignment has led to widespread adoption by major geospatial foundation models. SatMAE~\cite{Cong2022SatMAE}, developed at Stanford, was the first to adapt MAE for earth observation, introducing spatio-temporal masking on Sentinel-2 temporal stacks where 75\% of patches are masked in both space and time, forcing the model to learn temporal dynamics of Earth's surface. Trained on the fMoW dataset with multiple timesteps, SatMAE demonstrated that the MAE approach, originally designed for static natural images, transferred effectively to multispectral and multitemporal satellite imagery, achieving 71.77\% accuracy on land cover classification versus 54.46\% for a supervised baseline.

Additional MAE's concrete advantages manifest in three critical operational dimensions. First, the technical barrier to entry is low: implementing MAE requires few lines of code (take an open-source ViT via timm~\cite{Wightman2023rwightman} or Hugging Face, add random masking in 10 lines, a simple decoder in 20 lines, and train with MSE), versus 5000+ lines and often months of development for complex custom architectures. Second, the mature software infrastructure around the ViT/MAE ecosystem (Hugging Face transformers~\cite{Wolf2020Transformers}, timm, TensorFlow~\cite{tensorflow2015-whitepaper}/Keras~\cite{Chollet2015Keras} implementations, public pre-trained checkpoints, abundant tutorials) facilitates reuse and reproducibility. Finally, academic momentum follows clear path dependency: MAE accumulates 3000+ citations, SatMAE demonstrates earth observation feasibility, Prithvi~\cite{Jakubik2023Foundation-arxiv} brings institutional validation with NASA/IBM partnership and 2023-2024 unlocked an explosion of variants (Clay, ScaleMAE, etc.), creating a virtuous circle where each success reinforces adoption.

Indeed, Prithvi, jointly developed by NASA and IBM, represents the first major institutional validation of the MAE approach for earth observation. Based on a ViT-Large with 307M parameters, Prithvi is trained on 1 million HLS (Harmonized Landsat-Sentinel) 6-band images covering the continental United States for 2017, strictly following the MAE protocol with 75\% spatial masking. Results show substantial improvements over standard ViT-base baseline: +10.06pp\footnote{pp means \textit{percentage points}.} $\scoreFOne$ on flood detection (Sen1Floods11~\cite{Bonafilia2020Sen1Floods11}) reaching 90.71\% vs 80.65\%, and +2.64pp mIoU on fire scar mapping (84.84\% vs 82.2\%). However, when comparing pre-trained versus non-pre-trained Prithvi with identical architecture, the gains are modest: only +0.59pp IoU on flood detection (81.26\% vs. 80.67\%) and +1.36pp on fire scar mapping (73.62\% vs 72.26\%) after 500 epochs. The primary benefit of pre-training lies in convergence speed and data efficiency, where the pre-trained model achieves comparable performance with significantly fewer labeled samples (\eg maintaining >80\% IoU with only 126 images instead of 252 on flood detection, representing a 50\% reduction in labeling requirements). The main contribution thus lies in dramatically reducing annotation costs rather than improving final performance.

On the other hand, Clay~\cite{Clay2024TheFoundationModel} pushes the MAE approach further by introducing metadata conditioning in the architecture. Trained on 70 million global samples (70$\times$ more than Prithvi), totaling 33.8 TB of data, Clay supports multiple heterogeneous sensors (Sentinel-1 radar, Sentinel-2 optical, Landsat) by conditioning the ViT encoder on metadata including spatial resolution, sensor type, wavelengths, and timestamp. This sensor-aware approach enables the model to learn sensor-invariant representations while capturing each modality's specifics. The model achieves a mIoU of 0.4819~\cite{Wiratama2025Comparative} on land cover classification (11 classes) in Indonesia, surpassing U-Net++ (0.4073) and Prithvi (0.3678). Clay particularly excels on complex classes such as general plantations (0.7789 IoU), clouds (0.8296), and canals (0.4661). From this point on, multimodality gave rise to a new class of models.

\subsection{The multimodal shift}

Two recent models particularly illustrate this transformation: Prithvi-EO-2.0~\cite{Szwarcman2026PrithviEO2.0} from IBM/NASA in December 2024 and TerraMind~\cite{Jakubik2025TerraMind} from IBM/ESA/NASA in February 2025. They represent an evolution from the pure MAE approach of Prithvi-EO-1.0 (2023, 100M parameters) toward fundamentally more sophisticated systems capable of integrating heterogeneous modalities across different spatial and temporal scales.

Prithvi-EO-2.0/-TL~\cite{Szwarcman2026PrithviEO2.0} is trained on 4.2 million global samples of harmonized Landsat-Sentinel time series at 30m resolution with 300M and 600M parameter versions and represents the direct evolution of Prithvi-EO-1.0 by retaining the standard MAE approach while introducing explicit temporal and geographic embeddings into its ViT architecture. The model uses 3D patches (spatiotemporal) with t=1 to accommodate irregular satellite acquisition, and integrates geolocation (latitude/longitude) and temporal metadata (year, day of year) via sin/cos encodings added to tokens through learned weighted summation, including a random dropout mechanism (probability 0.1) during pre-training to handle potential absence of this information at inference. This multi-temporal approach with sequences of 4 timestamps spaced 1 to 6 months apart enables the model to learn seasonality and long-term trends while capturing short-term changes, surpassing its predecessor by 8\% on GEO-Bench~\cite{Lacoste2023GEOBench} and achieving state-of-the-art performance on 12 diverse tasks covering resolutions from 0.1m to 30m (including high-resolution RGB tasks like cattle or individual tree detection despite 30m pre-training). Results indicate that versions with temporal and geographic embeddings (Prithvi-EO-2.0-600M-TL) systematically outperform their equivalents without this metadata, validating the importance of explicitly contextualizing satellite observations in space and time.

TerraMind~\cite{Jakubik2025TerraMind} published two months after Prithvi-EO-2.0 by an overlapping team (J. Jakubik as the first author on both), represents an architectural shift rather than a simple iteration. 
The architecture introduces the first any-to-any generative model for Earth Observation with a dual-scale approach combining token-level and pixel-level representations simultaneously. 
It is trained on TerraMesh, a dataset of 9 million global samples at 10m resolution (more than the double of Prithvi-EO-2.0) integrating 9 modalities (Sentinel-1 GRD/RTC radar, Sentinel-2 L1C/L2A/RGB optical, DEM elevation, LULC 9 classes, NDVI vegetation, textual captions, geographic coordinates). TerraMind uses custom auto-encoder tokenizers with Finite Scalar Quantization~\cite{Mentzer2024Finite} (FSQ, 16K vocabulary) and diffusion decoders for each image-like modality, achieving compression rates up to 3,000$\times$ for Sentinel-2 L2A (12 bands).

To go deeper technically, the symmetric transformer-based encoder-decoder architecture (Base 480M and Large $>$600M parameter versions) simultaneously accepts pixel-level inputs (raw sensor data as floats) and token-level inputs (compressed discrete representations) for each 16$\times$16 patch, enabling learning of both cross-modal correlations via tokens (high-level abstraction) and fine spatial nuances via pixels (local details), thus structuring the embedding space better than token-only approaches. Furthermore, the masking during pre-training uses Dirichlet sampling to randomly select input and target tokens, forcing the model to adapt to entire missing modalities (crucial for earth observation where sensors fail).

TerraMind also introduces ``Thinking-in-Modalities'' (TiM), a recursive fine-tuning and inference technique inspired by chain-of-thought in LLMs where the model generates additional synthetic modalities (\eg LULC~\cite{Sawant2023Sen2} from Sentinel-2) that are concatenated with the original input to enrich the input space, improving performance by +2.49pp IoU on water mapping (Sen1Floods11) with S-2+generated LULC versus S-2 alone, and up to +4.25pp IoU with S-1+generated LULC. On the PANGAEA~\cite{Marsocci2024PANGAEA-arxiv} benchmark (9 datasets), TerraMindv1-L achieves 59.57\% mean mIoU (rank 3.44), becoming the first earth observation foundation model to systematically outperform task-specific U-Nets trained from scratch (+1pp mIoU in unimodal, +4pp in multimodal, peaking at +4.5pp on multimodal tasks).

In 2025, the AlphaEarth Foundations Model~\cite{Brown2025AlphaEarth-arxiv} (AEF) explicitly positions itself as a critical evolution beyond MAE-based approaches, acknowledging its lineage while identifying its limitations for demanding operational applications. The AEF paper states: ``\textit{A new generation of geospatial foundation model approaches can be roughly characterized as derivatives of SatMAE [...] while these approaches represent progress [...] none satisfy all of the following key properties: (1) multi-source or multi-modality, (2) inclusion of time into the modeling framework, or (3) spatial resolution at a precision useful for serving operational mapping use cases.}''. The model deviates significantly from the standard MAE paradigm across several dimensions. 

It is important to note that no decisive breakthrough has truly emerged. In 2024, PhilEO Bench~\cite{Fibaek2024PhilEO} revealed that foundation models such as SatMAE and Prithvi are outperformed by a simple U-Net with skip connections on image-to-image tasks for building detection. This finding highlights the critical role of well-designed benchmarks. Harder, closer to real applications, and selective enough to reveal genuine differences between approaches.

\subsection{The case of AlphaEarth}

AEF is a geospatial embedding model developed by Google DeepMind that generates universal representations of Earth's surface at 10m resolution with compact 64-dimensional vectors (64 bytes per pixel). An embedding is a compact, dense representation that transforms complex, high-dimensional data into a reduced-dimensional vector space while preserving important semantic properties.
AEF has been trained on 3 billion images from 9 heterogeneous sources (Sentinel-1/2, Landsat, radar, LiDAR, climate, gravity, topography, etc.) and uses a new Space-Time-Precision (STP) architecture with 480M parameters that explicitly models time as a continuous variable, enabling temporal interpolation and extrapolation. The fundamental principle is to model dynamics, not just static states. It does not learn to memorize static images but to model the underlying temporal processes that generate observations, similar to how one can predict a planet's position without watching it at every instant by knowing its orbital trajectory.

For example, given temperature data for the city of Ispra in Italy, for January, February, March and then September, October, November, a human would interpolate the June temperature based on seasonal patterns to estimate around 22 degrees for a summer month. AEF learns these patterns across millions of locations. During training, AEF is forced to develop this capability through temporal masking strategies: forecasting (predicting the future), backcasting (reconstructing the past), or masking images from certain seasons that the model must learn to reconstruct correctly. The model learns local rates of change, seasonal patterns, development trends, and how urban terrain textures evolve.
Its truly key innovation lies in a ``teacher-student'' system that guarantees identical embeddings even with 30-50\% missing data  through multi-objective learning combining multi-source reconstruction and temporal coherence. Unlike classical decoders like MAE, AEF has implicit functions that can generate any instant. The decoder takes as input a 64D embedding summarizing the period $[t\_s, t\_e)$, in a normalized time representing position format in $[0, 1)$,  indicating which moment is desired, sensor metadata containing acquisition geometry (angle, orbit, etc.), and can learn to generate N frames for N instants.

The model reduces errors by 23.9\% on average (1.4$\times$ improvement) compared to the best existing approaches across 15 evaluation tasks (classification, regression, change detection, etc.), while requiring 16$\times$ less storage than competing methods. For post-disaster damaged building detection, AEF offers strong robustness to clouds (radar penetrates clouds), sensor failures (trained with dropout of entire sources), and enables unsupervised change detection through simple distance between embeddings before and after events. Note that global annual embeddings for 2017-2024 are freely accessible on Google Earth Engine, which enable immediate deployment without retraining for various applications with minimal data.

There are some limitations to note for building damage assessment direct-use, though. To use it straight, the 10m spatial resolution is still quite coarse at building scale. A 10m$\times$10m pixel still covers a significant area, and details like a single collapsed building or partial roof collapse can be lost. Among highlighted limitations is performance in extreme low-data regimes (few-shot). When labeled training data quantity is extremely low, AEF's gains become less clear and highly variable, which is often the case immediately after major disasters. The paper directly states: ``\textit{For 1000-fold 1-shot trials, AEF error reductions were only $>$ 1.0$\times$ in the ~90\% confidence interval in 5/15 evaluations with average variation ±0.49$\times$[...] we consider the extreme degree of variability indicative that adequate general 1-shot or 10-shot performance remains an unsolved research frontier.}''.
Temporal dependency also exists. The embedding is a temporal summary. If the period is too long, a sudden change like earthquake-induced collapse can be ``smoothed out'' and lost. The concept of ``valid period'' introduced by the paper here is also important. The model produces an embedding summarizing a period, not a precise instant. This impacts building damage detection: to detect sudden damage events (earthquake, explosion), comparing a very short ``before'' period with a very short ``after'' period is crucial. If the summary period is too long (\eg one year), the signal of abrupt change will be diluted in the embedding by numerous unchanged days, making detection much more difficult.

For example, in practice for change detection, a method could be to calculate two embeddings: one for a ``before'' period (\eg 2022) and one for an ``after'' period (\eg 2023) for the same pixel, then compute a similarity measure between the two 64D vectors. This value (between 0 and 1) calculated for each pixel represents change magnitude. Alternatively, another one could use a simple classifier by identifying dozens or hundreds of exemplar pixels for each class (\eg 50 ``destroyed building'' pixels, 100 ``non-damaged building'' pixels, 50 ``partially-damaged building'' pixels), extract embeddings to obtain 64D vectors for each training pixel, and train a classifier using simple algorithms (like logistic regression or k-nearest-neighbors). This trained classifier can then be applied to all pixels in the area of interest, predicting a class for each pixel and generating a visual classification map (\eg destroyed buildings in red, non-damaged buildings in green, as shown in \cref{fig:graphical_abstract}).

Low-shot learning results on the LUCAS Land Cover dataset containing the ``buildings\_with\_1\_to\_3\_floors'' class show balanced accuracy around 35\% across all 40 dataset classes. Recognizing 40 landscape types with only 300 examples each, tested on 3629 pixels, represents honorable performance.

\subsection{Perspectives}

Beyond AEF and MAE, several alternative paradigms emerge in the earth observation foundation model landscape. SatCLIP~\cite{Klemmer2025SatCLIP} adopts a contrastive approach inspired by original CLIP, aligning location encoders (geographic coordinates) and image encoders via InfoNCE loss without any pixel reconstruction. This approach produces 256-dimensional embeddings learning geospatial semantics directly through contrast rather than reconstruction, with the advantage of not requiring expensive decoders and producing high-quality global embeddings optimized for semantic similarity. However, SatCLIP suffers from typical contrastive learning limitations: requiring very large batch sizes (2048-4096) for sufficient negatives causing high memory consumption, sensitivity to augmentation choices (problematic for defining relevant augmentations across 13 spectral bands), and absence of dense spatial structure making fine segmentation more difficult. Results nonetheless show competitive performance, particularly on global classification tasks like US trees (39 genera), where SatCLIP excels through learning geographic relationships via its location encoder.
Diffusion models also represent an emerging frontier (2023-2024), with works like DiffusionSat~\cite{Khanna2024DiffusionSat} and SatDM~\cite{Baghirli2023SatDM-arxiv} adapting diffusion models to earth observation. Instead of masking and reconstructing like MAE, these models learn to progressively denoise images to which Gaussian noise has been iteratively added, a process simultaneously producing generative capabilities (synthesis of realistic satellite data) and rich representations for discriminative tasks. Yet, these approaches have potentially prohibitive computational costs (iterative denoising is intrinsically slower than single forward inference of MAE) and open questions about training stability and scalability to tens of millions of samples.

Novel approaches like AEF represent another direction, combining reconstruction (MAE-like), contrastive learning (CLIP-like), consistency learning (teacher-student), and realistic structural masking in a unified multi-objective framework. These architectures offer a good generalization by capturing different aspects of data structure, superior operational robustness to real-world conditions, and multi-task learning benefits where objectives mutually regularize each other. The cost is increased architectural and algorithmic complexity, often longer development times, and more difficult reproducibility as subtle implementation details can strongly impact results. But for critical applications that need to be super reliable (like disaster monitoring, real-time change detection, and global operational mapping), these trade-offs can be worth it for the performance and reliability gains.

\newpage
\section{Organizing a challenge for building damage assessment}
\label{sec:challenge}

The previous sections have highlighted the lack of consensus on the task formulation, the data, and the evaluation of \bdaAcronym methods. 
One proven way to make progress on all three fronts simultaneously is to organize a \textit{challenge}: a common task, a common dataset, and a common evaluation protocol, applied to all participating methods. 
This section examines how such a challenge could be set up. 
We first revisit the data aspects that condition a challenge (\cref{sec:data_3}), then the evaluation strategy (\cref{sec:evaluation_3}), and finally the organizational aspects and scenarios (\cref{sec:challenge_3}).

\subsection{Data: a key component for the evaluation of data-driven method}
\label{sec:data_3}

In \cref{sec:data}, we discussed the different types of data used for the \bda task and the discrepancies between data annotation conventions. These points are revisited below from the perspective of organizing a challenge, together with the case for a private dataset.

\subsubsection{Data type and annotations}

As seen in \cref{sec:data}, there are several types of data sources, primarily characterized by their modality and resolution, as well as several annotation schemas.

\paragraph{Raw data.} 
Regarding the raw data, the issue of image accessibility is paramount. Indeed, high-resolution images are not available to the public or the scientific community\footnote{It is however possible to submit a request to the ESA to obtain high-resolution historical images.}. Furthermore, given that high-resolution images pose difficult-to-resolve problems due to the angle of inclination of the captures ---which introduces perspective effects that are difficult to interpret in the case of collapsed buildings--- it seems reasonable to favor, in a scientific challenge, the use of Sentinel-2-type images for the \bdaAcronym task\footnote{Experience shows that Sentinel-1 SAR imagery alone is not sufficient to solve the task of \bdaAcronym at the building level. Note that the BRIGHT dataset~\cite{Chen2025Bright} relies on VHR commercial SAR (Capella/Umbra, 0.3 to 1 m), which suffers less from the same resolution limitation but raises the availability and cost issues discussed above, making it less suitable for the targeted challenge.}. %
High-resolution images would then be reserved for critical service applications. 
It should nevertheless be noted that the issue of image availability makes the matter more delicate than it appears. Indeed, Sentinel-2 images are unusable in the presence of disturbances such as clouds. One must then either wait for the next pass (which is not acceptable for an emergency service) or fall back on other satellites. Finally, to obtain images, one must commission an acquisition mission. In other words, satellites of private companies only carry out a mission if they are commissioned for a specific acquisition, which entails a cost, with no guarantee that the acquisition will be of sufficient quality. 

In any case, we advocate having a set of \textit{homogeneous data (single modality of images and fixed resolution) for scientific exploration}, given that the results of a remote sensing method depend on the resolution used~\cite{Bou2025ChangeDetection}, but \textit{heterogeneous data for assessing the possibility of practical deployment to mimic real-world constraints}.

\paragraph{Annotations.} 
Regarding annotations, we explained in \cref{sec:data_annotations} that there are different types of annotations and several classification typologies (see \cref{tab:annotations-summary}).
From a technical standpoint, the type of annotations (pixel-wise, bounding boxes, points) is irrelevant as long as the definition is unambiguous. 
The issue of typology is more critical. Indeed, the CEMS ODM~/~IWG-SEM~\cite{IWG-SEM2018Emergency} concept of ``Possibly damaged'' is potentially questionable because it is a source of ambiguity during the annotation process. 

\paragraph{Digital Terrain Model ({DTM})\index{DTM}.} 
To find a solution, it is also necessary to understand how the process of identifying damaged buildings is carried out. 
In practice, developers have access to digital terrain models (DTMs), which represent the ``bare earth'' surface, with all above-ground objects removed, leaving only the underlying terrain; in lidar processing, this corresponds to ground returns. 

There is, for example, the  FABDEM\index{FABDEM} (Forest And Buildings removed Copernicus Digital Elevation Model DEM\index{DEM}) model. It is a global bare-earth DTM derived from the Copernicus DEM by algorithmically removing the signal contributions of vegetation canopy and built structures. FABDEM uses machine learning corrections to produce a cleaner bare-earth surface, making it much closer to a true DTM at global scale. There are also the DSMs (Digital Surface Model)\index{DSM} and derived DTMs from the company named Maxar (formerly DigitalGlobe); Maxar produces very high-resolution DSMs and derived DTMs using stereo photogrammetry from their WorldView satellite constellation (WorldView-1 through -4), which can achieve sub-meter spatial resolution ---far finer than freely available global products like SRTM or the Copernicus DEM (both $\sim30\ m$). 
Anyone can obtain the FABDEM DTMs, but the resolution is insufficient for the task since it is $\sim30\ m$.

There exists an alternative to the use of a DTM. Indeed, there are cadastral data (public and non-public), and the \citetitle{OpenStreetMap} service (\url{https://www.openstreetmap.org}) that provide the ground-level outlines of buildings. This information is very useful for solving the \bdaAcronym task. This means that we can locate the majority of buildings in a disaster-stricken area. 
This has two consequences: (1) during a challenge, participants can submit both the dates and the DSM provided by OpenStreetMap or a partial DSM, and (2) a new class could be added to the annotation typology.

\paragraph{A new classification typology for the task of \bdaAcronym.} 
We therefore propose the alternative typology for the classification task, as described in \cref{tab:annotations-new}. Using the localization enabled by OpenStreetMap or any other building-footprint source, we define the ``potentially\footnote{We avoid the term ``Possibly'' to prevent any confusion with the previous terminology used in \cite{IWG-SEM2018Emergency}.} damaged'' class to indicate a building near which we detect either debris or another building that is destroyed or damaged. This approach allows us to account for cases where a building has collapsed or is damaged (\eg, cracks) without this being visible from an aerial view because, for example, the roof appears intact.

\begin{table}[t]
    \centering
    \caption{A possible new nomenclature for the task of \bdaAcronym.}
    \begin{tabular}{c}
    \toprule
            Destroyed\\
    \midrule
            Damaged\\
    \midrule
            Potentially damaged (visible damages or debris in the surroundings)\\
    \midrule
            No visible damage (undamaged)\\
    \midrule
            Background\\
     \bottomrule
    \end{tabular}
    \label{tab:annotations-new}
\end{table} 

Beyond the issue of typology, the use of a DSM also affects the format of annotations. Indeed, the simplest annotation involves clicking on a point (placing a marker) within a defined area on the DSM. The task then becomes ``\textit{instance-based segmentation}''.

The issue of the ``background'' class is not as straightforward as it seems. In fact, the detected clouds are unlabeled areas. The question then becomes whether cloud areas should also be provided, given that there are several cloud detection algorithms and various types of clouds. By providing common cloud masks during a challenge, we can avoid any bias.

Finally, one last question concerns the grade of the detected damages and the associated confidence level. Ideally, we would like to be able to define the damage rate on a scale from $0$ to $100\%$, while also displaying a confidence score. The results of a challenge could show how far we still have to go; therefore, organizers could suggest participants provide a damage rate and a confidence score, and invite them to see if they can link it to a crisp instance segmentation. 

There is, of course, a cost associated with this new classification system since existing data must be re-annotated, and adding a class always increases the cost of annotation. On the other hand, it is always preferable to have an unambiguous class label.

\subsubsection{Data split}

The integrity of the evaluation is a cornerstone of progress in solving artificial intelligence (AI) tasks. A common difficulty is how the data is organized (data split). Another recurring issue is that once test data is published, it becomes accessible to model developers and may end up being used in training processes, which inflates reported performance scores and compromises scientific reproducibility. This effect has been observed with the development of transformers, which are trained almost exclusively on all publicly available annotated data on the web, accounting for much more than the provided dataset and potentially suffering from leaking knowledge.

When defining a challenge, it is therefore necessary to distinguish between data used for training and developing new methods, challenge data, and any undisclosed data. 
Below we discuss how the data split contributes to building trust between a developer and a user, and we expand our discussion to include the concept of a golden sample.

\begin{figure}[th]
    \centering
    \includegraphics[width=0.6\textwidth,alt={Levels of trust between a developer and a user for three setups.}]{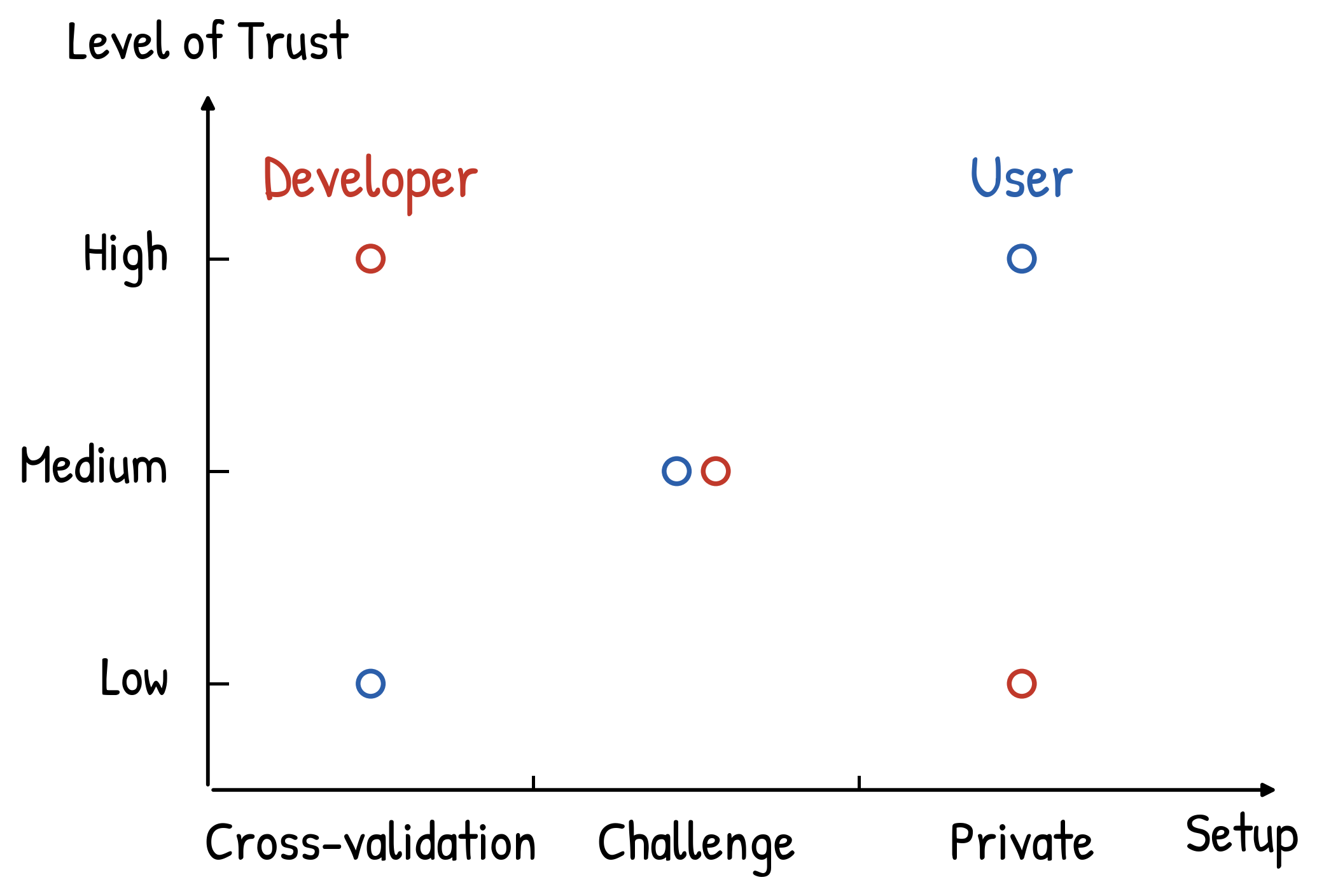}
    \caption{Levels of trust between a developer and a user for three setups.}
    \label{fig:pillars-trust-setup2}
\end{figure}

\paragraph{Building trust between a developer and a user is essential.} 
\Cref{fig:pillars-trust-setup2} 
provides an overview of the trust levels between a developer and a user based on three different setups.

In the \textbf{cross-validation setup}, the developer’s practice involves dividing the data and annotations into three groups: (1) the training set, (2) the validation set, and (3) the test set. 
The separation of data into a training set and a disjoint test set is a foundational concept in statistical learning theory~\cite{Vapnik1995TheNature}; \citet{Hastie2009TheElements} noted that mixing these roles ---\eg, choosing the model with the best test performance--- is a form of information leakage. 

The purpose of this separation is to obtain an unbiased estimate of generalization performance; the test set acts as a proxy for unseen production data. 
The development of the method is then performed on the training set. The validation set serves as a safeguard to check for the risk of overfitting. As for the test set, it is intended to evaluate the method’s performance as if it were a real-world situation. In theory, the test set should be used only once. An assumption that is rarely followed in practice, bringing some form of light overfitting.
Since developers work in a perfectly controlled environment, they will have a high degree of confidence in the method they produce and the results obtained. In contrast, the user will place very little trust in the method. 
When organizing a challenge, there is little point in requiring splits for training data since (1) it is impossible to verify compliance with the splits and (2) evaluation is based solely on the challenge set. The only question is whether to provide additional raw data without annotations. This may seem trivial, but giving certain participants access to non-public data (which may be annotated) could skew the results of a challenge. Unfortunately, it is particularly difficult to find a fully satisfactory answer to this question other than asking participants whether they used additional data.

In the case of the \textbf{challenge setup}, the raw data is shared but not the annotations. This is a tradeoff setup in which the developer will have reasonable confidence since he/she knows the data on which their method will be tested, and the user can verify the quality of the results thanks to the annotations that only they know. 
In \cref{sec:challenge_3}, we will explain various scenarios for organizing challenges.

The third possibility is that of the \textbf{private setup}, which is similar to the practice of using the ``holdout'' set in the xBD challenge (see \cref{sec:method-Literature}). In this case, as shown in \cref{tab:setup_availability2}, the developer has no access to data and annotations. 
As a result, the user is certain that a method does not overfit the data or annotations typical for their application. Confidence in the results of a method is very high. On the other hand, for the developer, it is quite the opposite. Indeed, since the developer is unaware of the data to which their method is applied, they will have low confidence in the results obtained by applying their method, as indicated in \cref{tab:setup_availability2}. 
\begin{table}[th]
    \caption{What is known from the developer for the different setups.}
    \medskip
    \centering
    \begin{tabular}{lcc}
          \textbf{Setup}&\textbf{Data is known}&\textbf{Annotations are known} \\ 
          \hline
          Cross-validation & Yes & Yes \\
          Challenge & Yes & No \\
          Private & No & No
    \end{tabular}
    \label{tab:setup_availability2}
\end{table}

This setup requires the use of samples that we will refer to as ``golden samples''. This point is discussed in the following section.

\subsubsection{Private setup and golden samples}
\label{sec:golden_sample_main}

Evaluating a machine-learning model requires data that was not available during training. This principle, as old as statistical learning theory itself, is routinely violated in modern AI research, either unintentionally ---through the uncontrolled crawling of large web corpora--- or deliberately, through \emph{benchmark gaming}.

In industrial practice, the term ``golden sample'' is used to describe samples not shared with the developer, although the term ``golden sample'' is not explicitly defined in scientific literature or in standardization. 
Below, we define the notion of golden sample and focus on how it can be used.

A \textbf{golden sample} is a test instance (or a collection of such instances, sometimes called a \emph{golden set} or \emph{private holdout}) that satisfies the following properties simultaneously:

\begin{enumerate}
  \item \textbf{Non-disclosure.}  The sample is never released publicly, not even after evaluation; its content, statistics, and any derivative information that could enable reconstruction are kept secret to avoid any kind of data leakage. 
  \item \textbf{Evaluative purpose.} The sample is used to produce a performance estimate that might be reported to developers and the public, but without exposing the sample itself.
  \item \textbf{Exclusivity of custody.} A golden sample is held exclusively by an independent evaluating party, that might be a regulatory body, a competition organizer, or a scientific consortium, and is not shared with any developer or participant.
\end{enumerate}

This combination distinguishes golden samples from the ordinary test split of a publicly released dataset, from validation sets shared with participants, and from withheld labels for the purpose of a challenge.  
The key insight is that \emph{any exposure ---even indirect, such as a numerical score returned after
each submission--- degrades the golden property over time} because scores can be used as an oracle to infer the content of the private set.

The central concern with private data and golden samples is whether an entity has the capacity to collect golden samples, both logistically and financially. Indeed, these samples ---which will not be disclosed--- must be annotated, which in some cases does nothing to build trust between a developer and a user.

\subsection{Towards an evaluation strategy}
\label{sec:evaluation_3}

Before examining the organizational aspects of challenges, it is important to note that evaluation remains the central issue.

\subsubsection{Indirect leakage through score reporting}

 The most widely used evaluation method is the public leaderboard, popularized by machine-learning competition platforms such as Kaggle (founded in 2010). Kaggle splits the test set into a \emph{public} portion used to compute a live score during the competition and a \emph{private} portion, revealed only
at the end.  The private portion is a practical example of a golden sample: participants see their public score, but not their private score, and the private test data itself is never released.  This design was introduced precisely to prevent ``overfitting the leaderboard'', \ie using repeated
submissions to the scoring oracle to reconstruct the test labels~\cite{Recht2019DoImagenet}.

Even when the test data itself is never released, repeated reporting of scores on the same fixed test set constitutes a slow information leak.  \citet{Blum2015TheLadder} showed that this adaptive process can, over time, allow a developer to overfit to a private test set without ever seeing it. The ARC Prize organisers acknowledged this risk explicitly: by 2024, approximately 10,000 scores had been reported against the same 100-task private evaluation set, representing a non-trivial risk of overfitting~\cite{Chollet2024ARCPrize-arxiv}.

\subsubsection{The need for a revised evaluation strategy}

In the \bdaAcronym community, the central issue regarding the evaluation strategy is the choice of a score. Since there is no consensus within the community and annotations differ, it is necessary to review the evaluation strategy. 

The following excerpt, taken from the article of \fullcite{Bolelli2026Multistructure},  speaks for itself and perfectly illustrates the difficulties encountered in certain challenges.

\begin{quote}
\textbf{Ranking}. Although following standard and consolidated approaches (Maier-Hein et al., 2018), a key limitation of our evaluation protocol is that the final ordering is obtained by aggregating ranks across multiple tasks/targets, which makes the leaderboard sensitive to the number and type of submissions, particularly when the same group submits multiple times (Piérard et al., 2025). From a theoretical perspective, performance-based rankings should ideally satisfy consistency properties (\eg the relative order between two methods should not change simply because other methods enter or leave the pool). To reduce instability, we considered only the latest submission per team. However, despite enforcing this rule on the Grand Challenge platform, we later detected likely circumvention via multiple accounts linked to the same group. These cases were evident because they yielded identical scores across many classes and test cases: an implausible outcome for independently trained models. We contacted the accounts for clarification, but received no response; therefore, we removed all duplicate submissions from the final-test leaderboard before computing the final ranking.

Future editions of the challenge will strengthen (i) team/account verification and submission policies, (ii) automatic detection of near-duplicate submissions, and (iii) rank-stability audits (\eg leave-one-out sensitivity). Methodologically, we will also investigate ranking procedures more closely aligned with the axiomatic view of performance-based ranking (\eg preorder-based formulations and explicit ``importance'' weighting) to reduce dependence on participant pool composition (\citet{Pierard2025Foundations}).
\end{quote}

Beyond the issue of attempts to manipulate ranking scores, it is worth noting that the authors emphasize the stability of scores relative to the ranking, which represents a fairly radical departure from standard ranking approaches. One example among many is the article \cite{Krasnodebska2025Advancing}, in which the authors justify the preference for a modified Jaccard coefficient\footnote{The Jaccard coefficient is presented as an extension of the IoU coefficient.} over a modified $F_1$-type score. 
This is perfectly justifiable from a practical standpoint during use (provided one fully understands the selected score), but this choice is not justified when comparing methods since it is known that the Jaccard coefficient and the $F_1$ score produce the same ranking ~\cite{Pierard2025Foundations}. In other words, \emph{it is the stability of the ranking, and not the absolute value of a score, that should take precedence when comparing methods}.

\subsubsection{The impossibility of predicting performance on a new dataset}
\label{sec:impossibility}

One aspect that remains underestimated by the artificial intelligence community is the difficulty of producing results on new data that are similar to those obtained during a test. This difficulty is highlighted, for example, in the following two articles:  

\begin{itemize}
    \item In the abstract of  \citetitle{Dehghani2021TheBenchmark-arxiv} (\citet{Dehghani2021TheBenchmark-arxiv}), we read  
    \begin{quote}
    This paper proposes the notion of ``a benchmark lottery'' that describes the overall fragility of the machine learning benchmarking process. The benchmark lottery postulates that many factors, other than fundamental algorithmic superiority, may lead to a method being perceived as superior.
\end{quote}
    \item There is a similar conclusion in \citetitle{Pierard2025AMethodology} (\citet{Pierard2025AMethodology}), which concludes that
    \begin{quote} 
    In the case that we studied, we have shown that the performance-based ranking of computer vision methods on new domains is far from being an easy or solved problem. 
    \end{quote}
\end{itemize}

The inability to predict or guarantee future performance means that: 
\begin{itemize}
    \item We must remain realistic in our ambitions. This likely also means \emph{accepting a certain margin of error or tolerance during benchmarking.} 
    \item \emph{This inability may justify the use of golden samples}, as it is a way to reduce the risk relative to expectations. But we have noticed that this is not always possible. 
\end{itemize}

\paragraph{Averaging scores: `mind the gap'.} 
In addition to the difficulty of predicting performance on new data, there is the challenge of defining an average score. Several types of averaging are widely discussed in the literature: arithmetic mean, geometric mean, and harmonic mean. In reality, only the \emph{contextual mean}, defined in \cite{Pierard2020Summarizing}, is compatible with a probabilistic framework such as that required to define precision, recall, and the $F_1$ score. In the case of an unconditional score, the contextual average is the arithmetic mean, but for conditional scores, the contextual average incorporates priors. In practice, this means that the solution for averaging performance is to use the contextual average. As an example, a complete probabilistic performance evaluation pipeline, compatible with averaging, has been proposed recently by \citet{Zakharova2026Event-arxiv} for the detection of events in videos.

\subsubsection{Practical lessons}

In summary, here are a few recommendations regarding the evaluation: 

\begin{enumerate}

\item Make sure to \emph{define/express a task properly.} Unfortunately, it is not uncommon for the task to be defined incorrectly, which consequently complicates the process of interpreting the results.

\item \emph{Interpreting a single score is debatable} as long as the score is not closely related to the task interpretation. Instead, for general purposes, prefer
\begin{itemize}
\item using multiple scores (see the Tile~\cite{Pierard2024TheTile-arxiv} for example).
\item ranking/comparing a new method to a baseline. This approach allows us to begin ranking methods and avoids focusing on the absolute value of a score.
\end{itemize}

\item \emph{Use the contextual average for averaging}. Averaging scores blindly with the arithmetic mean is unsound in most cases. 

\item \emph{Predicting a performance on new data is not working well}. We must accept that evaluating a model on challenge data alone will not be enough to guarantee its performance on new, unseen data. This may also mean that multiple methods are used to solve a problem, and that the choice of method depends on the location. 
For example, using Marie Tharp’s map~\cite{TharpHeezen1957Atlantic} or the Global Earthquake Model (GEM) Foundation’s Global Active Faults dataset~\cite{Styron2020TheGEM} showing the locations of seismic faults, one could describe an area by identifying the typical type of housing found in a high-risk region.

\item \emph{There is a need for new benchmarking strategies}, which might involve never sharing private data or mixing the results on different evaluation sets. 
\end{enumerate}

\subsection{Organization of a challenge}
\label{sec:challenge_3}

In this section, we will take a look at the practical aspects on how to organize a challenge.

\paragraph{Computer vision challenges.} 
The PASCAL Visual Object Classes (VOC) challenge~\cite{Everingham2010PascalVOC} introduced a two-phase design that became a template for computer vision competitions.
In the first phase, annotated training and validation sets are distributed. In the second phase, unannotated test images are released and participants submit predictions before a deadline; ground-truth labels are withheld until after submissions close and are then published.  This design provides temporary secrecy of labels during the competition but does not satisfy the strict golden-sample criterion because labels are eventually released.

The ImageNet Large Scale Visual Recognition Challenge
(ILSVRC)~\cite{Russakovsky2015ImageNet} followed the same pattern on a larger scale, with a scoring server that evaluated submissions against hidden labels. The test set labels were eventually released, enabling subsequent contamination of later models' training corpora.

\subsubsection{Limitations of the approach for building trust over time}

In the previous sections, we discussed the cases of challenge and private setups. 
Because of the amount of trust of both the developer and the user, it may seem that the challenge setup is the best configuration for fine-tuning methods. Unfortunately, things are not that simple if one wants a method to stand the test of time. Indeed, as the developer optimizes the parameters of a method to achieve the best results on the challenge set, one will eventually end up with the equivalent of annotations for the challenge set. From there, user confidence will erode since this amounts to overfitting the challenge set data. 

This means that the content of the challenge set must be regularly refreshed. An interesting approach is to first move the data from the private setup to the challenge setup and then to the cross-validation setup by incorporating the annotations. This way, one rebuilds the developer’s confidence. However, as the user exhausts their golden samples, they will no longer have private data and may lose confidence in the results of a method since everything will ultimately end up in the cross-validation setup.

\subsubsection{Technical possibilities}

After discussing the setups, let’s move on to the organization of the challenge itself. From a formal viewpoint, there are many paradigms to implement a challenge. We will analyze a few of them below.

\paragraph{Permanently withhold test sets with a scoring server.}

This is the simplest paradigm: the evaluation data are stored on a server controlled by an independent party.  Participants submit predictions; the server returns a scalar score. The data are never transmitted to participants.
\textsc{CodaLab}~\cite{Pavao2023CodaLab}, Codabench~\cite{Xu2022Codabench}, and \textsc{EvalAI}~\cite{Yadav2019EvalAI-arxiv} provide open-source platforms that implement this pattern with configurable submission quotas and leaderboard policies.

\textbf{Limitations.}  As \citet{Blum2015TheLadder} show, if the number of allowed submissions is not tightly controlled, the scoring oracle leaks information adaptively. The rate of leakage is proportional to the number of submissions, making unlimited submission policies dangerous. With $n$ test samples and $s$ submissions, up to $O(s / \sqrt{n})$ units of information may be extracted. In practice, the number of submissions allowed per correctly identified participant is always limited to prevent manipulation of the leaderboard.

\subsubsection{The sealed exam paradigm}

\citet{Cruz2026LLMOlympiad-arxiv} propose an \emph{Olympiad-style} evaluation event inspired by academic examination design. Tasks are sealed until the evaluation date. Participant systems are frozen in advance and all entries are
run through a single standardized evaluation harness controlled by an independent party.  
After scoring, the full task set and evaluation code are
released for auditability. This design makes strong performance harder to manufacture ---a model cannot be fine-tuned on the tasks because they were not known in advance--- while restoring reproducibility post hoc.

The sealed-exam paradigm can be seen as a time-limited golden sample: the test data are golden during the competition window and public afterward.  
The post-publication contamination of future models is accepted as a deliberate trade-off in favor of auditability. 
The sealed exam paradigm is a suitable candidate for the implementation of a challenge for the task of \bdaAcronym.

\subsubsection{Three types of scenarios}
\label{sec:challenge_three_scenarios}

In the following, we will distinguish between different challenge setups, which are justified by different objectives and result in different organizational structures. These challenge types (setups) are defined as follows: 
\begin{description}
    \item[S1: Scientific scenario (with unannotated test data).] 

    In this setup, the objective is to develop the method that best meets the performance expectations. Transparency is essential to issuing a scientific opinion. In short, participants in this type of challenge seek both to publicize their work to secure funding and to establish their scientific expertise through publications.

    In this scenario, the test set (without labels) is released to participants during the challenge period.  Participants submit predictions before a deadline. Labels are withheld during the challenge but are released after the deadline (or at least after all submissions are scored). 

    \item[Call for tender (CfT) scenario.] 
    
    This involves responding to a call for proposals issued to the market; it is the setup considered when one wishes to develop a service based on the use of a remote sensing method. Two parties are involved: the bidder, motivated by the prospect of being selected, and the contract awardee, whose goal is to obtain a service that meets their specific needs.

    \begin{itemize}
        \item \textbf{S2: CfT \textit{with} golden samples.}
        A purchasing body (typically a government agency) evaluates competing vendors on data that none of them have ever seen. Results inform a procurement decision, and the evaluation data are never released
        to any vendor, before or after the decision.  
        
        \item \textbf{S3: CfT \textit{without} golden samples.}
        Vendors are evaluated on publicly available benchmarks, or on validation data provided to them in advance. The evaluating body uses published leaderboard scores or scores provided by the vendor itself as the basis for selection. This is the de facto practice in most enterprise AI procurement today.
    \end{itemize}

\end{description}

We can see, therefore, that the configuration of these scenarios differs considerably; this is why we compare them in the next section. 
\Cref{tab:comparison} compares these three scenarios across several evaluation-integrity dimensions.

\paragraph{Analysis.} 
Scenario~S1 (scientific challenge) represents a well-established practice that prioritizes \emph{reproducibility and community learning}: by releasing both images and eventual annotations, the challenge contributes a reusable dataset to the research community.  The cost is that labels eventually enter the public domain and can contaminate  training corpora of future models.  
This is an accepted trade-off in academic settings but is unacceptable in high-stakes commercial or regulatory settings.

Scenario~S2 (CfT with golden samples) provides the strongest evaluation guarantee: scores are trustworthy, gaming is nearly impossible, and no information leaks to future competitors. The cost is low reproducibility:
independent researchers cannot verify the scores, and the test distribution is opaque. 
This trade-off is appropriate when the \emph{decision validity} of a single procurement or regulatory approval is more important than long-term scientific progress.

Scenario~S3 (CfT without golden samples) is the most dangerous from an integrity standpoint. In practice, many enterprise AI procurement decisions today fall into this category --vendors are evaluated on their self-reported
benchmark scores or on demonstration data they have contributed to themselves. This creates strong incentives for Goodhart-Law-style gaming~\cite{Manheim2018Categorizing-arxiv}, which can be quoted as \textit{when a measure becomes a target, it ceases to be a good measure}.

The scenarios are not mutually exclusive.  A well-designed challenge can combine elements of S1 and S2 by running two parallel evaluation streams: a \emph{public scientific stream} (S1, annotations released post-challenge) for academic progress and reproducibility, and a \emph{CfT commercial stream} (S2, golden samples never released) for procurement or regulatory decisions. ARC-AGI-2~\cite{Chollet2026ARCPrize-arxiv} already
approximates this hybrid design.

\begin{table}[p]
\centering
\caption{Comparison of three evaluation scenarios across evaluation-integrity dimensions.
$\checkmark$~=~property holds;
$(\checkmark)$~=~partially holds;
$\times$~=~property does not hold;
?~=~depends on implementation;
N/A~=~not applicable.}
\label{tab:comparison}
\small
\begin{tabular}{p{3cm}>{\centering\arraybackslash}p{3.5cm}>{\centering\arraybackslash}p{3.5cm}>{\centering\arraybackslash}p{3.5cm}}
\toprule
\textbf{Dimension}
  & \textbf{S1: Scientific scenario}
  & \textbf{S2: CfT scenario}
  & \textbf{S3: CfT scenario} \\
  & \textbf{(unannotated test)}
  & \textbf{(with golden samples)}
  & \textbf{(no golden samples)} \\
\midrule
Test data released to participants
  & $\checkmark$ (without labels)
  & $\times$
  & $\checkmark$ or $(\checkmark)$ \\
\midrule
Annotations / labels released
  & After deadline
  & Never
  & Before or during evaluation \\
\midrule
Independent evaluator custody
  & $(\checkmark)$ (by organiser)
  & $\checkmark$ (strict)
  & $\times$ \\
\midrule
Training contamination risk
  & Low during challenge; high post-release
  & Very low
  & High \\
\midrule
Deliberate benchmark gaming risk
  & Medium (label guessing possible)
  & Very low
  & High \\
\midrule
Reproducibility of results
  & High (post-challenge)
  & Low (secret data)
  & Medium \\
\midrule
Auditability by third parties
  & $\checkmark$ (post-release)
  & $(\checkmark)$ (by custodian only)
  & $(\checkmark)$ \\
\midrule
Longitudinal comparability
  & Good (stable test set until release)
  & Limited (set may be retired)
  & Good (public benchmarks stable) \\
\midrule
Trustworthiness of scores
  & High (during window)
  & Highest
  & Low--medium \\
\midrule
Participant fairness
  & $\checkmark$ (same data for all)
  & $\checkmark$ (same data for all)
  & ? (depends on prior exposure) \\
\midrule
Post-event annotation release
  & $\checkmark$ (supports future research)
  & $\times$ (never released)
  & N/A \\
\midrule
Protects future models from contamination
  & $\times$ (labels released eventually)
  & $\checkmark$
  & $\times$ \\

\bottomrule
\end{tabular}
\end{table}

\subsubsection{Summary}

The first question to ask is what the primary goal of launching a challenge is. There are several points to consider:
\begin{enumerate}
 \item Do we want to evaluate or compare methods? These are two entirely different tasks! 
 \item What is the purpose of benchmarking? Do we want to select the best method? To stimulate the scientific community. Overall, we need a ``problem statement'' for the benchmarking.
 \item How do we plan to use the results? Basically, what are they for? We can allow for some flexibility, particularly by using a weighting variable, as is done in \cite{Pierard2025Foundations}.
\end{enumerate}

\subsubsection{Practical aspects}

Similar to the SoccerNet challenges~\cite{Giancola2022SoccerNet,Cioppa2024SoccerNet2023Challenge, Giancola2025SoccerNet-arxiv}, 
organizing a \bdaAcronym challenge requires careful planning across several dimensions.

First, an \textbf{annotated dataset} must be created, which is typically the most costly operation in terms of human and financial resources. The data must also faithfully reflect the objectives of the methods that participants are expected to develop.

Second, \textbf{open-source code} should be provided to lower the barrier to entry. This includes a data loader to handle input/output operations, at least a simple baseline method to serve as a reference point for performance comparison and as a starting point for newcomers, and evaluation tools to avoid any ambiguity in score calculation.

Third, an \textbf{infrastructure for independent performance evaluation} must be set up. This typically relies on online platforms such as EvalAI (now a paid service) or Codabench (free of charge), where participants can submit their predictions and consult their results, and optionally those of other participants. We recommend keeping the leaderboard private for the duration of the challenge to encourage broader participation, as early high-performing submissions may discourage newcomers from entering.

Fourth, \textbf{keeping the challenge open to diverse profiles} is important. Participants from different backgrounds (academia, industry, and independent researchers) bring complementary perspectives and fresh ideas. Even lower-performing methods can yield valuable insights into promising or ineffective approaches to the task.

Fifth, \textbf{disseminating the results} through a challenge report co-authored with the participants is essential. This serves both as a motivation for participation and as a mechanism to ensure a successful rollout for subsequent editions of the challenge. It also contributes to keeping the broader scientific community engaged and active.

Finally, \textbf{promoting participation} through dedicated channels, such as a website, a Discord server, and YouTube videos, is important. A challenge is only as alive as its community, and sustaining it requires answering questions, being responsive to inquiries, and actively recruiting participants over the years. This is not a one-time effort but rather a long-term commitment, with the ultimate goal of building a lasting community around the task.

\newpage
\section{Conclusion}
\label{sec:conclusion}

In this chapter, we revisited the task of \bda (\bdaAcronym) by satellite imaging along three pillars: the data, the evaluation, and the  methods.

Our starting point was the observation that the task itself is defined in multiple, often inconsistent, ways across the literature. 
Depending on the work, \bdaAcronym is cast as a binary classification, a multi-class classification, a semantic segmentation, or an instance-level detection problem. 
These formulations are not interchangeable: they imply different annotations, different methods, and different evaluation protocols, and the frequent confusion between them undermines the comparability of published results. 
We argued that an operational task definition should be data-agnostic, tolerate missing pre-event imagery, and support heterogeneous inputs, and that instance-based formulations offer a good compromise between localization, annotation feasibility, and operational interpretability.

Concerning the data pillar, we analyzed the available datasets along three axes, namely the modality, the temporality, and the geographic scope, and we examined annotation practices and their pitfalls: ambiguous class typologies (in particular the ``possibly damaged'' class), misalignments and geolocation errors, and incomplete coverage caused by the time-constrained nature of rapid mapping. 
Concerning the evaluation pillar, we showed that current practices, which are largely based on a single score such as the $F_1$ score, are theoretically fragile, and we advocated an evaluation philosophy based on rankings, multiple scores, distinct application cases, and sound multi-domain strategies.
Concerning the methods pillar, we reviewed the evolution from task-specific deep networks to Earth observation foundation models, noting that, despite their promise, no decisive breakthrough has yet emerged for \bdaAcronym, and that generalization to unseen events and regions remains the fundamental difficulty. 

Finally, we discussed how these elements can be brought together in the organization of a challenge.
The distinction between the cross-validation, challenge, and private data-sharing setups, together with the notion of golden samples, provides a principled way to build trust between developers and users over time. 
We outlined concrete scenarios, from a scientific challenge aimed at engaging the research community to a call for tender aimed at selecting an operational service provider, and we described the practical requirements for running such a challenge successfully.

We hope that this chapter contributes to establishing \bdaAcronym on firmer scientific ground, by encouraging precise task statements, transparent data and annotation practices, and theoretically sound evaluation protocols, so that future methods can be developed, compared, and deployed with confidence when disasters strike.

\newpage
\printbibliography

\newpage
\printindex{}

\newpage

\begin{thenomenclature} 
\nomgroup{A}
  \item [{AOI}]\begingroup Acronym for \textit{area of interest}\index{AOI}. It is the geographic area where the emergency or mapping analysis is carried out, and it can be a polygon, rectangle, circle, or other defined shape.\nomeqref {0}\nompageref{4}
  \item [{AUROC}]\begingroup The \textit{Area Under the Receiver Operating Characteristic (AUROC or AUC)}\index{AUROC} is a single-number summary of a binary classifier’s ability to discriminate between the positive and negative classes. It is the probability that a randomly chosen positive instance is ranked higher (\ie, assigned a higher predicted score) than a randomly chosen negative instance by the classifier. An AUROC equal to 0.5 implies no better discrimination than random chance.\nomeqref {0}\nompageref{4}
  \item [{BDA}]\begingroup Acronym for \textit{building damage assessment}.\nomeqref {0}\nompageref{4}
  \item [{Binary}]\begingroup A \textit{binary} classification task is a supervised machine learning problem where each input must be assigned to one of exactly two possible classes, such as positive vs. negative or 0 vs. 1. It is also called a crisp two-class classification task.\nomeqref {0}\nompageref{4}
  \item [{Classification}]\begingroup A \textit{classification}\index{classification} task in machine learning is a predictive problem where the goal is to assign each input example to one label from a predefined set of categories.\nomeqref {0}\nompageref{4}
  \item [{Cross-validation}]\begingroup \textit{Cross-validation}\index{cross-validation} is a way to evaluate how well a model will perform on new, unseen data by repeatedly training and testing it on different splits of the same dataset. It is commonly used in machine learning to get a more reliable estimate of performance and to reduce overfitting.\nomeqref {0}\nompageref{4}
  \item [{Data splitting}]\begingroup \textit{Data splitting}\index{split} in machine learning involves dividing a dataset into distinct subsets, typically training\index{training}, validation\index{validation}, and test\index{test} sets, to train models and evaluate their performance on unseen data. This process ensures models generalize beyond the training data rather than overfitting.\nomeqref {0}\nompageref{4}
  \item [{DEM}]\begingroup A \textit{Digital Elevation Model (DEM)}\index{DEM} represents the bare-earth surface, removing all natural (trees for example) and built features (building, bridges, etc.). It represents the true topographical ground surface using data, typically from LiDAR or satellite imagery, processed to remove above-ground objects.\nomeqref {0}\nompageref{4}
  \item [{Detection}]\begingroup A \textit{detection}\index{detection} task in machine learning is a prediction problem where the goal is to decide whether one or more target patterns are present and, often, where they occur in the input data. It focuses on identifying the existence of specific events, objects, or anomalies rather than assigning every input to a fixed set of classes. From a more formal point of view, detection differs from binary classification in that detection does not involve true negatives.\nomeqref {0}\nompageref{4}
  \item [{DSM}]\begingroup A \textit{A Digital Surface Model (DSM)} captures both the natural and built/artificial features of the environment.\nomeqref {0}\nompageref{4}
  \item [{DTM}]\begingroup A \textit{Digital Terrain Model (DTM)}\index{DTM} in remote sensing is a digital representation of the terrain or ground surface of a specific area or location. It augments a a DEM in the sense that it provides information about the elevation, slope, and other topographic features of the terrain.\nomeqref {0}\nompageref{4}
  \item [{EMS}]\begingroup EMS\index{EMS} stands for Copernicus \textit{emergency management service}. It is an EU Earth-observation service that provides geospatial information and maps to support response to natural disasters, man-made emergencies, and humanitarian crises.\nomeqref {0}\nompageref{4}
  \item [{FABDEM}]\begingroup FABDEM\index{FABDEM} stands for ``Forest And Buildings removed Copernicus DEM''. It is a global bare-earth DTM derived from the Copernicus DEM. Developed at the University of Bristol and published around 2022, FABDEM addresses a well-known limitation of widely used global elevation products like SRTM and the Copernicus DEM: because those are DSMs, they systematically overestimate terrain elevation in forested and urban areas, which introduces significant errors in flood modeling, sea-level rise assessments, and hydrological routing. FABDEM uses machine learning corrections to produce a cleaner bare-earth surface, making it much closer to a true DTM at global scale.\nomeqref {0}\nompageref{4}
  \item [{Foundation model}]\begingroup A \textit{foundation model}\index{foundation model} is a large-scale model pretrained on broad data in a task-agnostic way (usually by self-supervised learning) that can be adapted to many downstream tasks via fine-tuning or prompting; large language and multimodal models are typical examples.\nomeqref {0}\nompageref{4}
  \item [{Generalization}]\begingroup \textit{Generalization}\index{generalization} in machine learning is the ability of a model to perform well on new, unseen data after being trained on a given dataset. It reflects how well the patterns learned from the training data apply to examples drawn from the same distribution as that data.\nomeqref {0}\nompageref{4}
  \item [{Imbalance}]\begingroup \index{imbalance}It refers to the distribution of classes in the input data. A dataset that contains the identical amount of samples per class is said to be balanced.\nomeqref {0}\nompageref{4}
  \item [{MAE}]\begingroup \index{MAE}Acronym for \textit{Masked Autoencoder}. It is a self-supervised approach for pretraining vision models in which a large fraction of image patches (typically ~75\%) is randomly masked and an encoder-decoder network is trained to reconstruct the missing pixels. The encoder processes only the visible patches, making pretraining efficient; the lightweight decoder is discarded afterward, leaving an encoder whose learned representations transfer well to downstream tasks.\nomeqref {0}\nompageref{4}
  \item [{mIoU}]\begingroup The \textit{Mean Intersection over Union}\index{mIoU} calculates the Intersection over Union (IoU) for each class---defined as the overlap area between predicted and ground truth segmentation masks divided by their union area---then averages these IoU scores across all classes. \nomeqref {0}\nompageref{4}
  \item [{Overfitting}]\begingroup \textit{Overfitting}\index{overfitting} in machine learning occurs when a model learns the training data too closely, including its noise and outliers, rather than capturing the underlying patterns. This results in excellent performance on training data but poor generalization to new, unseen data.\nomeqref {0}\nompageref{4}
  \item [{UAS}]\begingroup An \textit{Unmanned Aerial System} (UAS)\index{UAS} is the complete operational system that includes the UAV, ground control station, communications links, payloads, data links, software, and the human operator or ground crew needed to plan, monitor, and control flights.\nomeqref {0}\nompageref{4}
  \item [{UAV}]\begingroup \textit{UAV}\index{UAV} stands for unmanned aerial vehicle. UAVs can be terrestrial or aerial.\nomeqref {0}\nompageref{4}

\end{thenomenclature}
\end{document}